%% file: main.tex
\documentclass{article}

\PassOptionsToPackage{numbers, compress}{natbib}
\usepackage[a4paper, total={6in, 8in},top=30mm,bottom=30mm]{geometry}

\usepackage[utf8]{inputenc} 
\usepackage[T1]{fontenc}    
\usepackage{hyperref}       
\usepackage{url}            
\usepackage{booktabs}       
\usepackage{amsfonts}       
\usepackage{nicefrac}       
\usepackage{microtype}      
\usepackage[dvipsnames]{xcolor}

\usepackage[utf8]{inputenc}
\usepackage[T1]{fontenc}
\usepackage{hyperref}
\usepackage{url}
\usepackage{booktabs}
\usepackage{amsmath,amsfonts,amssymb,amsthm}
\usepackage{thmtools}
\usepackage{thm-restate}
\usepackage{nicefrac}
\usepackage{microtype}
\usepackage[inline]{enumitem}
\usepackage{graphicx}
\usepackage{multirow}
\usepackage{comment}
\usepackage{wrapfig}
\graphicspath{ {./figures/} }

\usepackage{mathtools}

\newcommand{\indep}{\perp \!\!\! \perp}
\newcommand{\real}{\mathbb{R}}
\newcommand{\norm}[1]{\left\lVert#1\right\rVert}
\newcommand\numberthis{\addtocounter{equation}{1}\tag{\theequation}}

\usepackage{cite}

\include{macros}

\title{Conditionally Invariant Representation Learning \\for Disentangling Cellular Heterogeneity}

\author{
Hananeh Aliee \\
\and
Ferdinand Kapl
\and
Soroor Hediyeh-Zadeh
\and
Fabian Theis
\and
\small Helmholtz Munich \\
\small Technical University of Munich\\
\small\texttt{\{hananeh.aliee,ferdinand.kapl,soroor.hediyehzadeh,fabian.theis\}}\\
\small\texttt{@helmholtz-munich.de}
}
\date{}

\begin{document}

\maketitle

\begin{abstract}
This paper presents a novel approach that leverages domain variability to learn representations that are conditionally invariant to unwanted variability or distractors.
Our approach identifies both spurious and invariant latent features necessary for achieving accurate reconstruction by placing distinct conditional priors on latent features.  
The invariant signals are disentangled from noise by enforcing independence which facilitates the construction of an interpretable model with a causal semantic. 
By exploiting the interplay between data domains and labels, our method simultaneously identifies invariant features and builds invariant predictors. 
We apply our method to grand biological challenges, such as data integration in single-cell genomics with the aim of capturing biological variations across datasets with many samples, obtained from different conditions or multiple laboratories. 
Our approach allows for the incorporation of specific biological mechanisms, including gene programs, disease states, or treatment conditions into the data integration process, bridging the gap between the theoretical assumptions and real biological applications.
Specifically, the proposed approach helps to disentangle biological signals from data biases that are unrelated to the target task or the causal explanation of interest.
Through extensive benchmarking using large-scale human hematopoiesis and human lung cancer data, we validate the superiority of our approach over existing methods and demonstrate that it can empower deeper insights into cellular heterogeneity and the identification of disease cell states.

\end{abstract}

\section{Introduction}


Learning high-level, latent variables from multi-domain\footnote{In this paper, \emph{multi-domain} data refers to different datasets that may have different characteristics, such as different distributions or different sources of biases.} data that explain the variation of data within each domain as well as similarities across domains is an important goal in machine learning~\cite{yin2021optimization, guo2022causal, sturma2023unpaired}.
The task involves identifying features of the training data which exhibit domain-varying spurious correlations that are unwanted and features which capture the true correlations of interest with labels that remain invariant across domains~\cite{lu2021nonlinear}.
It has been shown that the data representations or predictors that are based on the true correlations improve out-of-distribution generalization~\cite{ahuja2021empirical}.

The problem of learning invariant representations by exploiting the varying degrees of spurious correlations naturally present in training data has been addressed in several works~\cite{arjovsky2019invariant, kong2022partial}.
The general formulation of invariant representation learning is a challenging bi-level optimization problem, and existing theoretical guarantees often require linear constraints on data representations and/or classifiers~\cite{arjovsky2019invariant,ahuja2021empirical,ahuja2021invariance, ghassami2018multi}. 
The conditions and limitations of these approaches are extensively studied here \cite{rosenfeld2020risks, kamath2021does, guo2021out}.
In nonlinear settings, the work in~\cite{lu2021nonlinear} proposes a variational autoencoder (VAE) framework that uses a flexible conditionally
non-factorized prior to capture complicated dependencies between
the latent variables. 
It is a two-stage method that conducts the PC algorithm to discover the invariant latent variables. 
Despite the promising theory, the applications of invariant models for solving real-world problems, particularly in biology, is underexplored. 
%
%

Understanding the underlying factors and their dependencies within complex biological systems is a challenge in biology~\cite{rood2022impact}. 
In this paper, we focus on data integration and classification of multi-domain single-cell genomics data. 
Single-cell genomics allows us to study individual cells and their genetic makeup, providing a comprehensive view of the heterogeneity across cells.
However, datasets are often generated in different labs and under various experimental conditions where cells could have been exposed to chemical or genetic perturbations or sampled from individuals with diseases.
The exceptional complexity in such data makes the process of disentangling technical artifacts or distractors from relevant biological signals, referred to as data integration, non-trivial. 
Data integration in single-cell genomics is essential for capturing the cellular and molecular landscapes across domains. 
Deep generative models have shown great potentials in analyzing biological data~\cite{lopez2018deep}, but 
existing integration methods often struggle to distinguish relevant biological signals from noise, leading to over-correction and loss of biological variations. 

To address this challenge, we propose a conditionally invariant deep generative model that effectively integrates single-cell genomics data while preserving biological variations across datasets. 
Our model incorporates specific biological mechanisms, such as gene programs, disease states, or treatment conditions to capture the biological context of the data and facilitates deeper insights into cellular heterogeneity.
Our main contributions are as follows:
\begin{itemize}
    \item We revisit the fundamental assumptions in invariant representation learning and argue that in complex biological processes, the assumptions of independent and invariant causal mechanisms might not be sufficient.
    \item We propose an invariant representation learning method that identifies both spurious and invariant latent variables.
    \item We prove that our proposed method is identifiable up to a simple transformation and a permutation of the latent variables.
    \item We test our method using two large-scale datasets including human hematopoiesis~\cite{luecken2021a} and human lung cancer single-cell RNA-seq data~\cite{chan2021signatures} with 49 samples, spanning two lung cancer types and healthy individuals. 
    \item We intensively benchmark several invariant and identifiable deep generative models and demonstrate the superiority of our method for single-cell data integration, cell state identification, and cell type annotation.
\end{itemize}
\textbf{Related work.}
Invariant representation learning \cite{yin2021optimization,arjovsky2019invariant, lu2021nonlinear, pogodin2022efficient, shi2021invariant, peters2016causal, huang2020causal} leverages grouped
data from multiple environments for inferring the underlying invariant causal relationships, as learning the true causal structures between variables is not possible from i.i.d data from a single domain.
In traditional representation learning, independent component analysis (ICA)\cite{comon1994independent, hyvarinen2000independent} and non-linear ICA have been used to find independent latent features that generate the observations. However, identifying the true latent factors in the general nonlinear case of unsupervised representation learning is impossible \cite{khemakhem2020variational, lachapelle2022disentanglement, lopez2022learning, moran2021identifiable, hyvarinen1999nonlinear,locatello2019challenging, roeder2021linear}. 
A recent line of pioneering
works provide identifiability results for the nonlinear representation learning case under additional assumptions,
such as weakly- or self-supervised approaches which leverage additional information in the form of multiple
views \cite{gresele2020incomplete, locatello2020weakly, shu2019weakly, zimmermann2021contrastive, brehmer2022weakly}, auxiliary variables \cite{khemakhem2020variational,hyvarinen2019nonlinear, lu2021invariant} and temporal structure \cite{halva2020hidden, hyvarinen2016unsupervised, hyvarinen2017nonlinear, klindt2020towards, halva2021disentangling}.
To allow learning latent variables that may not be independent but causally related and motivated by \cite{scholkopf2022causality}, there has been a shift towards causal representation learning \cite{schölkopf2021causal, scholkopf2022statistical, gresele2021independent, leeb2020structural,yang2020causalvae,shen2022weakly,suter2019robustly, chevalley2022invariant,bengio2019meta, von2020towards}. Similar to works by \cite{khemakhem2020variational, lu2021nonlinear}, the present work is based on identifiable VAEs. However, unlike \cite{lu2021nonlinear}, our work disentangles latent representations that contain invariant information from the part that corresponds to changes across domains.

The current work is akin to recent works on causal disentanglement of style and content features in images with identifiability guarantees \cite{von2021self, kong2022partial, xiemulti, john2018disentangled, huang2022harnessing}, which partition the latent space into an invariant and a changing part. However, it differs from these works in the assumptions made on the data generation process, the causal graph and/or the requirement for having access to paired or multi-view observations \cite{von2021self}. Related work on the state-of-the-art generative models for single cell data is given in Appendix \ref{app:sec:integration}.


\section{Setup and background}
Consider an observed data variable $\textsc{x}^\textsc{u} \in \mathbb{R}^n$ from the domain $\textsc{u} \in \mathcal{U}$ with corresponding label $\textsc{y}^\textsc{u} \in \mathcal{Y}$ that is generated using a latent random vector $\textsc{z}^\textsc{u} \in \mathbb{R}^m$ (lower dimensional, $m\leq n$).
We are interested in learning a data representation $\Phi \in \mathcal{H}_{\Phi}: \mathcal{X}\rightarrow\mathcal{Z}$, where $\textsc{z}^\textsc{u}=(\textsc{z}_I,\textsc{z}_S)_{\textsc{z}_I,\textsc{z}_S \in\mathcal{H}_{\Phi}}$ and $\textsc{z}_I$ is invariant across domains $\mathcal{U}$. 
We can further use the latent representation $\textsc{z}_I$ for training an invariant predictor $\omega \circ \Phi$ such that $\omega \in\mathcal{H}_\omega:\mathcal{Z_I}\rightarrow \mathcal{Y}$ performs well across all domains, that is: 
\begin{equation}
\omega \in \text{argmin}_{\bar{\omega}\in \mathcal{H}_\omega} \mathcal{L}^\textsc{u}(\bar{\omega}\circ\Phi), \forall \textsc{u}\in \mathcal{U}
\end{equation}
where $\circ$ refers to function composition and $\mathcal{L}^\textsc{u}$ is the prediction loss in domain $\textsc{u}$.

Besides identifying the latent variables that are stable across environments, this paper is concerned with providing a causal semantic to those latent variables. 
Here, we briefly discuss some of the prominent related works.

\subsection{Identifiable VAEs}
The lack of identifiability in variational autoencoders (VAEs)~\cite{kingma2013auto} often leads to their failure in accurately approximating the true joint distribution of observed and latent variables. 
In the work presented in \cite{khemakhem2020variational}, the authors address this problem by extending nonlinear independent component analysis (ICA)~\cite{hyvarinen2016unsupervised,hyvarinen2017nonlinear,hyvarinen2019nonlinear} to a wide range of deep latent-variable models. 
They demonstrate that by employing a conditionally factorized prior distribution over observed and latent variables, denoted as $p_\theta(
\textsc{z}|\textsc{u})$, where $\textsc{u}$ represents an additional observed variable such as a class label $\textsc{y}$, identifiability can be ensured up to simple transformations. 

In other words, disregarding transformation for simplicity, if two different choices of model parameters $\theta$ and $\theta'$ yield the same marginal distribution $p_\theta(\textsc{x})=p_{\theta'}(\textsc{x})$, then it implies that $\theta = \theta'$, resulting in similar joint distributions $p_\theta(\textsc{x},\textsc{z}) = p_{\theta'}(\textsc{x},\textsc{z})$, and therefore similar posteriors $p_\theta(\textsc{z}|\textsc{x}) = p_{\theta'}(\textsc{z}|\textsc{x})$.

Let $\theta = (f, T, \lambda)$ represent the parameters of the conditional generative model as follows:
\begin{equation}
p_\theta(\textsc{x}, \textsc{z}|\textsc{u}) = p_f(\textsc{x}|\textsc{z})p_{T,\lambda}(\textsc{z}|\textsc{u})
\end{equation}
Here, $p_f(\textsc{x}|\textsc{z}) = p_\epsilon(\textsc{x} - f(\textsc{z}))$ where $f:\mathbb{R}^m\rightarrow\mathbb{R}^n$ is a non-linear injective function, and $\epsilon$ is an independent noise variable with a probability density function $p_\epsilon(\epsilon)$.
The identifiable VAE (iVAE)~\cite{khemakhem2020variational} assumes that the prior distribution $p_{T,\lambda}(\textsc{z}|\textsc{u})$ is conditionally factorial. 
In this assumption, each element $\textsc{z}_i \in \textsc{z}$ follows a univariate exponential family distribution characterized by the parameters $\lambda$ and the sufficient statistics $T$.

The identifiability result of iVAE is expanded upon in the work of non-factorized iVAE (NF-iVAE) by Lu et al.~\cite{lu2021nonlinear}. 
This extension relaxes the assumption of a factorized prior distribution for the latent representation, allowing for a more general exponential family distribution.
By adopting a non-factorized prior, the model gains the ability to capture complex dependencies between latent variables, which is often observed in real-world problems. 
However, the proposed model does not explicitly disentangle the latent variables that contain invariant information from the domain-specific variables that remain relevant for the prediction task.

\subsection{Invariant risk minimization}
The concept of invariant risk minimization (IRM)~\cite{arjovsky2019invariant} involves the development of an optimal and invariant predictor, denoted as $\omega$, that demonstrates strong performance across all environments, collectively referred to as $\mathcal{E}_{all}$.
The underlying assumption in IRM is that spurious correlations lack stability when observed across different environments ~\cite{woodward2005making}.
This assumption becomes particularly crucial when the training and testing datasets possess dissimilar distributions.
IRM utilizes training data gathered from distinct environments, represented as $\mathcal{E}_{tr}$ to learn a data representation $\Phi$ that enables the optimal classifier $\omega$ to consistently achieve satisfactory performance across all environments.
Mathematically, this concept can be formulated as follows:
\begin{equation}
    \begin{split} \min_{\substack{\Phi:\mathcal{X}\rightarrow\mathcal{H}\\ \omega:\mathcal{H}\rightarrow\mathcal{Y}} }
    \sum_{e\in\mathcal{E}_{tr}} R^e(\omega \circ \Phi)
    \end{split} 
\end{equation}
The authors in~\cite{arjovsky2019invariant} show that if both $\Phi$ and $\omega$ come from the class of linear models, under certain conditions, the predictor $\omega \circ \Phi$ remains invariant across $\mathcal{E}_{all}$.

\section{Problem specification}
Two commonly-made assumptions in causal representation learning are independent and invariant causal mechanisms which are generally seen as simplifying assumptions that may not hold across complex biological data.

The \emph{independent causal mechanisms (ICM)} principle \cite{scholkopf2012causal, peters2017elements} states that the generative process consists of independent modules that do not inform or influence each other. These modules are considered as conditional distributions of each variable in the system given its direct causes. 
%
%
Biological processes are, however, interconnected and can affect each other in complex ways. 
For example, genes do not work in isolation but instead interact with each other in the form of \emph{gene programs (modules)} or \emph{pathways} that carry out specific biological functions. 
Gene programs often share some genes and therefore they correlate. 
Additionally, there is a cross-talk between biological pathways, meaning that changes in one pathway or process can have downstream effects on other pathways and processes~\cite{subramanian2005gene,liberzon2011molecular}. 
There are, therefore, complicated dependencies between mechanisms in biological data where the ICM principle may not hold.


The \emph{invariant causal mechanisms} assumption is often made to enable generalization across different settings~\cite{mitrovic2020representation, chevalley2022invariant}. 
However, in biology invariance might not be an appropriate assumption since biological systems and processes can be affected by a variety of factors which can result in the causal relationships between variables to change over time or across different experiments.
For example, the causal relationship between a gene and a phenotype may depend on background genetic variation (e.g. mutations, cancer subtype, cell-line, treatment history in cancer patients, etc) that can simply vary across datasets generated from various patients under different conditions. 
In this case, an invariant causal learning model is very likely to leave out on important biological mechanisms that would be deterministic of patient outcome. 

To ensure generality in our model, we assume that latent factors can be dependent and refrain from interpreting inferred mechanisms as causally invariant unless there is sufficient evidence to support such an interpretation. 
Our focus instead is on separating noise and spurious correlations from biologically-interpretable factors within the model, with the aim of learning a structured representation that remains invariant to noise. 

\subsection{Motivating application in single-cell genomics}
Single-cell genomics allows us to study individual cells and their genetic makeup, providing a  comprehensive view of the heterogeneity across cells, cellular and molecular processes underlying biological systems~\cite{regev2017human}. 
However, analyzing single-cell data can be complex and challenging, as the datasets may be generated using different experimental techniques or platforms which add technical artifacts, called \emph{batch effects}, to the observations~\cite{lahnemann2020eleven}.
The datasets may also measure cells exposed to chemical or genetic perturbations \cite{dixit2016perturb}, or in diseases, where the presence of biological variations further complicates the analysis~\cite{lahnemann2020eleven}.

To address these challenges, \emph{data integration} methods aim to infer a representation of the datasets that captures the relevant biological signals while minimizing noise and technical artifacts. 
These methods help to identify the cell states that may be missed in individual datasets, 
helping us to build a comprehensive reference map, or \emph{atlas}, of the cellular and molecular landscapes. 
This reference map can then be used to compare and analyze new single-cell data~\cite{rood2022impact}. 


In the field of single-cell genomics, various methods have been proposed for data integration (see Appendix~\ref{app:sec:integration}). However, these methods often encounter challenges such as over-correction for batch effects and the removal of biological variations across similar cell states. 
Existing integration methods based on VAEs~\cite{lopez2018deep,xu2021probabilistic} also lack identifiability.
To address these limitations, our work introduces an identifiable deep generative model for effective data integration in single-cell genomics. 
This model disentangles representations that remain invariant across different environments from those that vary with the environment, such as batch effects. 
This disentanglement is crucial for preserving biological variations across datasets and surpassing the limitations of existing methods.
Moreover, our model can incorporate specific biological mechanisms, such as gene programs, disease states, or treatment conditions, allowing for their integration into the data integration process. 
This incorporation of biological factors enables the model to capture the biological context of the data and provide deeper insights into cellular heterogeneity.

\begin{figure}[!t]
\vspace{0cm}
\centering
\includegraphics[width=1\textwidth]{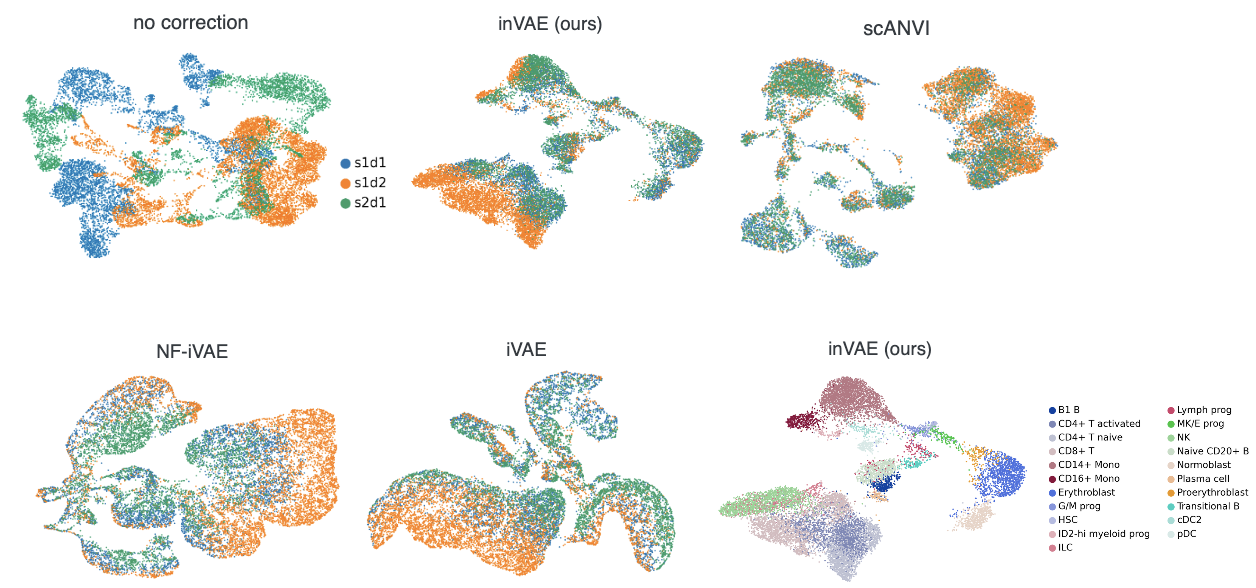}
\caption{Data integration using different methods. Shown are the UMAP representations of single cells colored by samples and cell types (bottom right). inVAE preserves biological variations between donors while mixing similar cells with technical noise (s1d1 and s2d1), whereas other methods either integrate out donor variability or do not merge cells from the same cell types.}\label{fig:motivation}
\end{figure}

\subsection{Assumptions on generative process}\label{sec:A1}
\begin{wrapfigure}{r}{0.45\textwidth}
\includegraphics[width=1.1\linewidth]{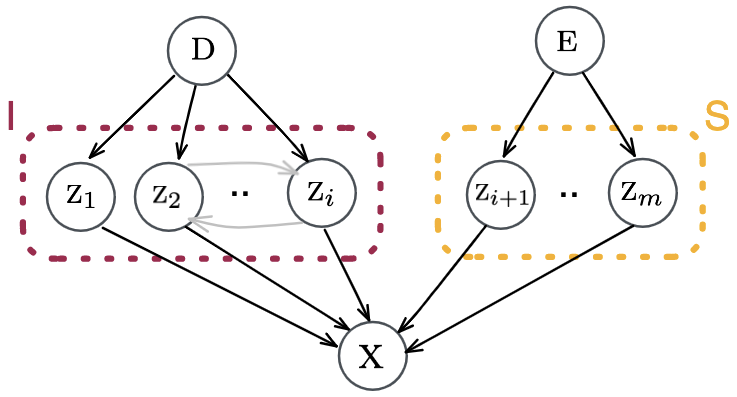} 
\caption{The data generating process.}
\label{fig:causal_graph}
\end{wrapfigure}
Our aim is to learn two sets of latent variables $\textsc{z}_I$ and $\textsc{z}_S$ such that $\textsc{z}=(\textsc{z}_I,\textsc{z}_S)$.
Latent variables $\textsc{z}_I \in \mathbb{R}^i$ represent inherent correlations of interest that remain consistent across different environments.
In contrast, latent variables $\textsc{z}_S \in \mathbb{R}^s$ (where $i+s = m$) correspond to spurious correlations that tend to fluctuate across varying environments.
We assume that the spurious correlations stemming from data biases are unrelated to the causal explanation of interest, although they are required for a perfect reconstruction.
We also substitute $\textsc{u}$ in iVAE as $(\textsc{d},\textsc{e})$ where $\textsc{d}$ refers to the specialized knowledge about the samples such as cell annotations, disease state, time, transcription factor, or gene program activity, etc. 
The variable $\textsc{e}$ also refers to the environment and encodes potential biases in the data. 

We assume that the causal graph underlying data generated under various conditions and environments satisfies the following assumptions:
\ass{\label{ass:causal_graph}(a) For any component $i$ in the latent space, $\textsc{z}_i$ depends \emph{either} on $\textsc{d}$ or $\textsc{e}~\ref{fig:causal_graph}$.
(b) The variables $\textsc{z}_I$ and $\textsc{z}_S$ are conditionally independent.
Therefore, $\textsc{z}_I$ is invariant to noise.
(c) The variable $\textsc{z}$ depends on both $\textsc{d}$ and $\textsc{e}$ and $p(\textsc{x}|\textsc{z})$ is not necessarily invariant across the environments.
(d) The label $\textsc{y}$ is independent of $\textsc{e}$ given $\textsc{z}_I$: $\textsc{y} \indep \textsc{e}|\textsc{z}_I$. Therefore, we assume $p(\textsc{y}|\textsc{z}_I)$ is invariant across all environments. 
}

The NF-iVAE model, as outlined in \cite{lu2021nonlinear}, operates under the assumption that the variable $\textsc{x}$ is conditionally independent of $\textsc{e}$ and $\textsc{d}$ given $\textsc{z}$: $\textsc{x} \indep \textsc{e},\textsc{d}|\textsc{z}$. 
This condition of independence suggests that the probability distribution $p(\textsc{x}|\textsc{z})$ remains consistent and invariant across all environments.
However, our findings, discussed in Section~\ref{sec:results}, reveal that in practical applications, environmental noise can still impact the latent representation $\textsc{z}$ of NF-iVAE, which challenges this assumption.
To tackle this issue, our proposed model disentangles spurious correlations $\textsc{z}_S$ and assumes that the resulting probability distribution $p(\textsc{y}|\textsc{z}_I)$ is independent of $\textsc{e}$ and, consequently, remains invariant across all environments.

We also assume that $\textsc{z}_I$ and $\textsc{z}_S$ are conditionally independent meaning that the invariant features are not biased and only capture the stable information across the environments:
\begin{equation}
p(\textsc{z}_I,\textsc{z}_S|\textsc{u}) = p(\textsc{z}_I|\textsc{u})p(\textsc{z}_S|\textsc{u}).
\end{equation}

\subsection{Assumptions on priors}
In accordance with Assumption~\ref{ass:causal_graph}, we make the following assumptions regarding the conditional priors to attain identifiability:

\ass{\label{ass:prior}(a) The conditional prior on the invariant data representation $p(\textsc{z}_I|\textsc{d})$ belongs to a general exponential family distribution that is not necessarily factorized. 
(b) Prior distribution $p(\textsc{z}_S|\textsc{e})$ is conditionally factorial.}

Assumption~\ref{ass:prior} (a) follows \cite{lu2021nonlinear} and allows for a more versatile and flexible prior given the domain specific knowledge that can effectively account for complex interdependencies between the invariant latent variables.
The conditional prior on the spurious latent variables $\textsc{z}_S$ can be either factorized or non-factorized depending on the application. 
In this paper, we assume that the spurious variables are factorized and independent of the invariant ones (Assumption~\ref{ass:prior} (b)).

The density function of the prior for the invariant latent variables $\textsc{z}_I \in \mathbb{R}^i$ is given by:
\begin{equation}
 \begin{split}
     p_{T,\lambda}(\textsc{z}_I|\textsc{d}) = \frac{\mathcal{Q}(\textsc{z}_I)}{\mathcal{Z}(\textsc{d})}\exp\left[ T(\textsc{z}_I)^T\lambda(\textsc{d})\right]
  \end{split}
  \label{eq:NF_prior}
\end{equation}
where $\mathcal{Q}$ is the base measure and $\mathcal{Z}$ is the normalizing constant (more details in Appendix~\ref{app:sec:training_loss}). 
The sufficient statistics $T(\textsc{z}_I) = [T_f(\textsc{z}_I)^T,T_{NN}(\textsc{z}_I)^T]^T$ are the concatenation of the sufficient statistics $T(\textsc{z}_I) = [T_1(\textsc{z}_{I_1})^T,..,T_n(\textsc{z}_{I_n})^T]$ of a factorized exponential family, where the dimension of $T_i(\textsc{z}_{I_i})$ is greater or equal to two.
$T_{NN}(\textsc{z}_I)$ is also the output of a neural network that allows the prior to model and capture arbitrary dependencies between the latent variables.

The density function of the prior for the spurious features $\textsc{z}_S \in \mathbb{R}^s$ is:
\begin{equation}
 \begin{split}
     p_{T,\lambda}(\textsc{z}_S|\textsc{e}) = \Pi_{i=1}^{s}\frac{\mathcal{Q}_i(\textsc{z}_{S_i})}{\mathcal{Z}_i(\textsc{e})}\exp\left[ \sum_{j=1}^{k}T_{i,j}(\textsc{z}_{S_i})\lambda_{i,j}(\textsc{e})\right]
  \end{split}
  \label{eq:F_prior}
\end{equation}
where $\textsc{z}_S$ are assumed to follow a factorized but unknown distribution $p(\textsc{z}_S) = \Pi_{i=1}^s p_i(\textsc{z}_{S_i})$~\cite{khemakhem2020variational}.
$\mathcal{Q}_i$ is the base measure, $\textsc{z}_{S_i}$ is the $i$-th dimension of $\textsc{z}_S$, $\mathcal{Z}_i(\textsc{e})$ the normalizing constant, $T_i = (T_{i,1},..,T_{i,k})$ the sufficient statistics, $\lambda_i(\textsc{e}) = (\lambda_{i,1},..,\lambda_{i,k})$ the corresponding natural parameters depending on $\textsc{e}$, and $k$ the fixed dimension of the sufficient statistics. 

\section{Proposed method}
\begin{figure}[t]
\centering
\includegraphics[width=1.0\textwidth]{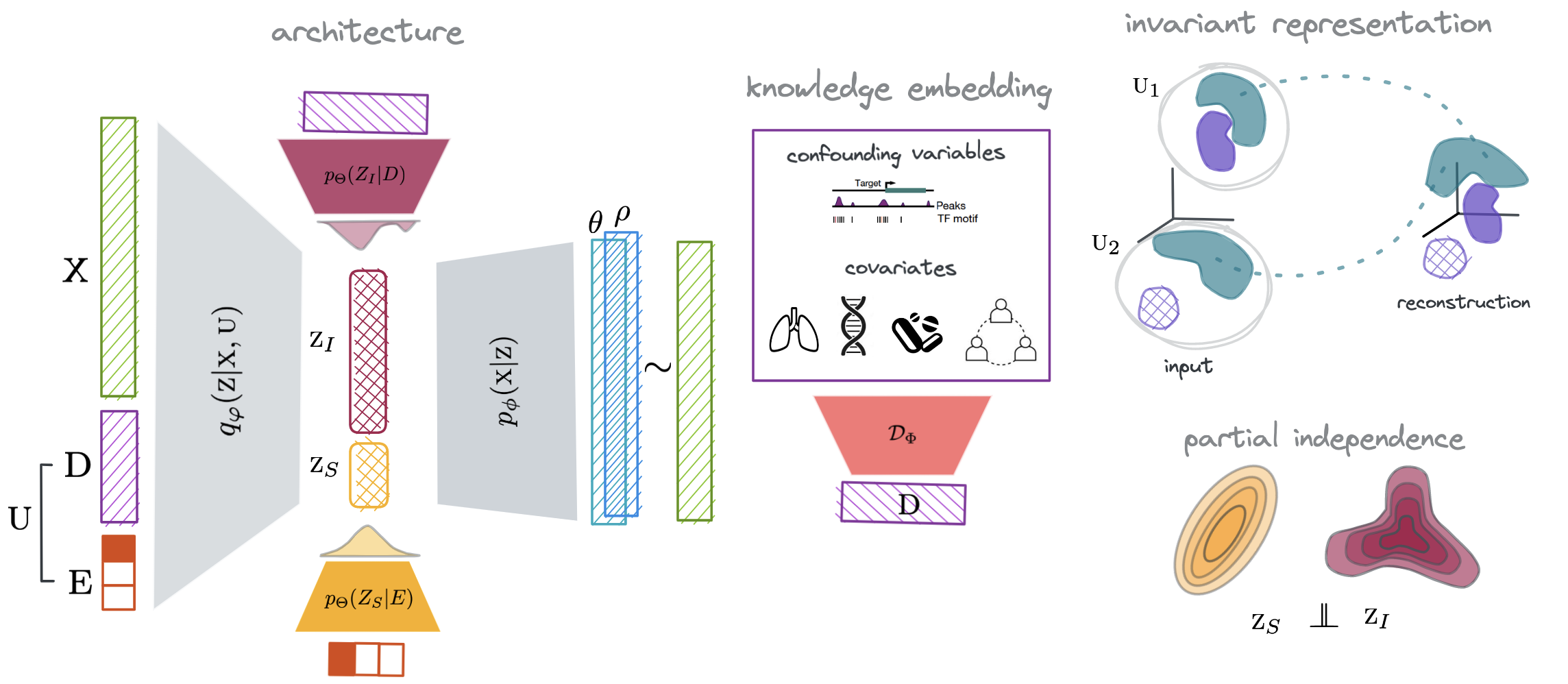}
\caption{Our VAE-based method. 
The observed variables $(\textsc{x},\textsc{d},\textsc{e})$ are encoded into two separate latent variables, namely $\textsc{z}_I$ and $\textsc{z}_S$. 
The variable $\textsc{e}$ is encoded using a one-hot encoder.
Prior knowledge, such as known biological covariates or mechanisms underlying the samples, is encoded through $\mathcal{D}_\psi$. 
The decoder component decodes the latent variables $\textsc{z}$ into the parameters of a {negative binomial distribution}, namely $\theta$ and $\rho$.
Two encoders infer the distribution parameters of the priors, denoted as $p_{\Theta}(\textsc{z}_I|\textsc{d})$ and $p_{\Theta}(\textsc{z}_S|\textsc{e})$. 
The proposed method learns a representation that is invariant to noise by disentangling $\textsc{z}_S$ and $\textsc{z}_I$.
}\label{fig:method}
\end{figure}
Our model, referred to as the invariant VAE (inVAE), is depicted in Figure~\ref{fig:method} and incorporates several key features. These features include: (i) disentangled spurious and invariant latent representations,
(ii) partially independent latent variables,
(iii) invariant representation using $\textsc{z}_I$,
(iv) identifiable data representation up to simple transformations.

The observed variables $(\textsc{x},\textsc{d},\textsc{e})$ are encoded into two separate latent variables, namely $\textsc{z}_I$ and $\textsc{z}_S$. 
The approximated posterior distribution, denoted as $q_\varphi(\textsc{z}|\textsc{x},\textsc{d},\textsc{e})$, represents the conditional distribution of the latent variables $\textsc{z}$ given the observed variables $(\textsc{x},\textsc{d},\textsc{e})$.
The variable $\textsc{e}$ is encoded using a one-hot encoder.
This component encodes the biases in the data.
Prior knowledge, such as known biological mechanisms  underlying the samples, is encoded through $\mathcal{D}_\psi$.
Two encoders infer the distribution parameters of the priors, denoted as $p_{\Theta}(\textsc{z}_I|\textsc{d})$ and $p_{\Theta}(\textsc{z}_S|\textsc{e})$, where $\Theta=(T,\lambda)$ for exponential distributions.
Finally, the likelihood $p_\phi(\textsc{x}|\textsc{z}_I,\textsc{z}_s)$ evaluates the reconstruction quality of the generated data.
In our VAE model, the decoder component decodes the latent variables $\textsc{z}$ into the parameters of a \emph{negative binomial distribution}. 
The negative binomial distribution is commonly used to model count data, which is often encountered in genomics applications. 

 \subsection{Loss function}
Our loss function contains three main parts: the evidence lower bound (ELBO) ($\mathcal{L}_{ELBO}$), the score matching (SM) loss ($\mathcal{L}_{SM}$), and the total correlation between the distributions of $\textsc{z}_I$ and $\textsc{z}_S$ ($\mathcal{L}_{TC}$).

The ELBO for the proposed model is as follows (proof in Appendix~\ref{app:sec:elbo:ours}):
\begin{equation}
 \begin{split}
     \mathcal{L}_{ELBO}(\varphi,\phi,T,\lambda):=& ~~\mathbb{E}_{p_D}\left[ \mathbb{E}_{q_\varphi(\textsc{z}|\textsc{x},\textsc{u})}\left[\log p_\phi(\textsc{x}|\textsc{z})\right] - D_{KL} (q_\varphi(\textsc{z}|\textsc{x},\textsc{u})||p_{T,\lambda}(\textsc{z}|\textsc{u}))\right] \\
     =&~~\mathbb{E}_{p_D}\left[\mathbb{E}_{q_\varphi(\textsc{z}|\textsc{x},\textsc{u})}\left[\log p_\phi(\textsc{x}|\textsc{z}) - \log q_\varphi(\textsc{z}|\textsc{x},\textsc{u}) + \log {p}_{T,\lambda}(\textsc{z}_I|\textsc{d}) + \log p_{T,\lambda}(\textsc{z}_S|\textsc{e}) \right] \right]
  \end{split}
\end{equation}
To model the factorized prior in Eq.~\ref{eq:F_prior}, we use Gaussian distribution in practice, so we can calculate the corresponding conditional probability by directly learning the mean and the variance of the distribution and optimize the ELBO in Eq.~\ref{app:sec:elbo:ours}, correspondingly.
However, the direct optimization of the non-factorized prior in Eq.~\ref{eq:NF_prior} is impossible as the normalization constant $\mathcal{Z}(\textsc{d})$ for a general multivariate exponential family distribution is unknown. 
In this work, we use score matching~\cite{hyvarinen2005estimation} for optimizing unnormalized probabilistic models, similar to the work in \cite{lu2021nonlinear}.
The score matching loss is as follows:
 \begin{align}
     \mathcal{L}_{SM} (T, \lambda) :&= -\mathbb{E}_{p_D}\left[\mathbb{E}_{q_\varphi(\textsc{z}_I|\textsc{x},\textsc{u})}\left[
	\norm{\nabla_\textsc{z} \log q_\varphi (\textsc{z}_I|\textsc{x},\textsc{u}) - \nabla_\textsc{z} \log \tilde p_{T, \lambda} (\textsc{z}_I|\textsc{d})}^2 \right]\right]  \label{eq:sm:original}\\
&= -\mathbb{E}_{p_D}\left[\mathbb{E}_{q_\varphi(\textsc{z}_I|\textsc{x},\textsc{u})}\left[
	\sum_{j=1}^i \left[ 
	\frac{\partial^2 \log \tilde{p}_{T, \lambda} (\textsc{z}_I|\textsc{d})}{\partial \textsc{z}_j^2} + 
	\frac{1}{2} \left( \frac{\partial \log \tilde{p}_{T, \lambda} (\textsc{z}_I|\textsc{d})}{\partial \textsc{z}_j} \right)^2\right]\right]\right] +cst. \label{eq:sm:practice}
  \end{align}
In practice to solve Eq.\ref{eq:sm:original}, we use partial integration to simplify the evaluation shown in Eq.\ref{eq:sm:practice} where $\tilde{p}_{T, \lambda}$ is the non-normalized density.
A summary of the score matching technique is provided in Appendix~\ref{app:sec:score}.

In this work, we aim at disentangling the noise and the invariant factors within the data. 
Previous studies, as suggested by Chen et al.~\cite{ChenCDHSSA16} and Higgins et al.~\cite{higgins2017betavae}, have highlighted the importance of two key aspects for achieving a disentangled representation: (i) maximizing the mutual information between the latent variables and the data variables, and (ii) promoting independence among the variables.
While the invariant variables may exhibit correlation in real-world applications, we only encourage independence between $\textsc{z}_I$ and $\textsc{z}_S$ by minimizing their total correlation. 
The total correlation is a measure of dependence between two random variables, and penalizing this correlation encourages the model to learn statistically independent factors~\cite{Watanabe1960InformationTA,NEURIPS2018_1ee3dfcd}.
To assess the total correlation between the distributions of $\textsc{z}_I$ and $\textsc{z}_S$, we employ the Kullback–Leibler (KL) divergence as follows:

\begin{align}\label{eq:TC}
    \mathcal{L}_{TC} :&= 
    D_{KL}\left(q_\varphi(\textsc{z}|\textsc{u})||q_\varphi(\textsc{z}_I|\textsc{u})q_\varphi(\textsc{z}_S|\textsc{u})\right)
\end{align}

Note that $q_\varphi(\textsc{z}|\textsc{u})$ requires the evaluation of the density 
$\mathbb{E}_{p(n|\textsc{u})}\left[ q_\varphi(\textsc{z}|\textsc{u},\textsc{x}_n)\right]$, where $p(n|\textsc{u})$ is the probability of observing the data point $\textsc{x}_n$ in the domain $\textsc{u}$.
We refer to $q_\varphi(\textsc{z}|\textsc{u})$ as the \emph{conditional aggregated posterior}, following \cite{MakhzaniSJG15}.
To measure the above density in practice, we extend the minibatch-weighted sampling proposed in~\cite{NEURIPS2018_1ee3dfcd} as follows:

\begin{equation}
    \mathbb{E}_{q_\varphi(\textsc{z}|\textsc{u})}\left[ \text{log}~q_\varphi(\textsc{z}|\textsc{u})\right]\approx 
    \frac{1}{b}\sum_{i=1}^{b} \left[\text{log}~\frac{1}{nb}\sum_{j=1}^{b}q_\varphi(\textsc{z}(\textsc{x}_i)|\textsc{x}_j,\textsc{u})\right]
\end{equation}
where $({\textsc{x}_1, ..., \textsc{x}_b})$ is a minibatch of samples and $\textsc{z}(\textsc{x}_i)$ is a sample from $q_\varphi(\textsc{z}|\textsc{u},\textsc{x}_i)$.
Both $\textsc{x}_i$ and $\textsc{x}_j$ are drawn from the same domain under the assumption of i.i.d data (proof in Appendix~\ref{app:sec:minibatch}).
We can similarly estimate $q_\varphi(\textsc{z}_I|\textsc{u})$ and $q_\varphi(\textsc{z}_S|\textsc{u})$ in Eq.~\ref{eq:TC}.

The overall loss is then:

\begin{equation}
  \begin{split}
  \mathcal{L}(\varphi,\phi, T, \lambda)
  =& ~~ \mathcal{L}_{ELBO} + \mathcal{L}_{SM} + \beta \cdot \mathcal{L}_{TC}\\
  =&~~\mathbb{E}_{p_D}\left[\mathbb{E}_{q_\varphi(\textsc{z}|\textsc{x},\textsc{u})}\left[\log p_\phi(\textsc{x}|\textsc{z}) - \log q_\varphi(\textsc{z}|\textsc{x},\textsc{u}) + \log \tilde{p}_{\hat{T},\hat\lambda}(\textsc{z}_I|\textsc{d}) + \log p_{T,\lambda}(\textsc{z}_S|\textsc{e}) \right] \right]  \\
  -&~~ \mathbb{E}_{p_D}\left[\mathbb{E}_{{q}_{\hat\varphi}(\textsc{z}_I|\textsc{x},\textsc{u})}\left[
	\sum_{j=1}^i \left[ 
	\frac{\partial^2 \log \tilde{p}_{T, \lambda} (\textsc{z}_I|\textsc{d})}{\partial \textsc{z}_j^2} + 
	\frac{1}{2} \left( \frac{\partial \log \tilde{p}_{T, \lambda} (\textsc{z}_I|\textsc{d})}{\partial \textsc{z}_j} \right)^2\right]\right]\right]  \\
  +&~~ \beta \cdot D_{KL}\left(q_\varphi(\textsc{z}|\textsc{u})||q_\varphi(\textsc{z}_I|\textsc{u})q_\varphi(\textsc{z}_S|\textsc{u})\right) \label{eq:training_loss}
  \end{split}
  \end{equation}
where $\beta$ is a tunable parameter and $\hat{\varphi},\hat{T},\hat{\lambda}$ are copies of ${\varphi},{T},{\lambda}$, and are treated as constant during the training.
The parameters of the encoder $q_\varphi(\textsc{z}|\textsc{x},\textsc{u})$, decoder $p_\phi(\textsc{x}|\textsc{z})$, and the {spurious} prior $p_{T,\lambda}(\textsc{z}_S|\textsc{e})$ are optimized using the $\mathcal{L}_{ELBO}$, while the invariant prior $\tilde{p}_{T,\lambda}(\textsc{z}_I|\textsc{d})$ remains fixed.
Additionally, the $\mathcal{L}_{SM}$ is employed solely for training the non-factorized prior with $\varphi$ being fixed.
Further implementation details are provided in Appendix~\ref{app:sec:training_loss}.

%
The identifiability proofs of each of the invariant and spurious representations as well as the joint representation are presented in~\ref{app:sec:identifiability}.

\subsection{Extension to discrete data with negative binomial distribution}

Negative Binomial is commonly used for modeling single-cell gene expressions~\cite{robinson2008small, hafemeister2019normalization}. 
Let $\rho_{cg}$ be the expected frequency of expression of gene $g$ in cell $c$ and $\theta_{g}$ be the inverse-dispersion of gene $g$. 
Denote $l_{c}$ as the library size (i.e. total read counts) in cell $c$. 
We use a non-linear transformation $f_{\phi}(\textsc{z})$ to generate the distributional parameters $\rho$ and $\theta$ of each gene in each cell. 
The gene expression $\textsc{x}_{cg}$ is then generated as $\textsc{x}_{cg}\sim \text{NegativeBinomial}(l_{c}\rho_{cg}, \theta_{g})$~\cite{robinson2008small, robinson2010edger, lopez2018deep}. 

\section{Results\label{sec:results}}
In this section, we present empirical results on two extensive datasets: human hematopoiesis~\cite{luecken2021a} and human lung cancer single-cell RNA-seq data~\cite{chan2021signatures}. 
We evaluate various deep generative models, including scVI \cite{lopez2018deep}, scANVI \cite{xu2021probabilistic}, iVAE \cite{khemakhem2020variational}, NF-iVAE \cite{lu2021nonlinear}, principal component analysis (PCA), and our own inVAE. 
We examine their impact on data integration, cell state identification, cell annotation, and generalization to unseen data—key challenges in single-cell genomics. 
We assess these methods using 7 different metrics outlined in~\cite{luecken2022benchmarking,scIB}. Appendix~\ref{app:sec:evaluation} provides an explanation of the metrics, and the architectures are discussed in Appendix~\ref{app:sec:arch}.

\subsection{Invariant representation of single-cell hematopoiesis data}
\begin{figure}[!t]
\vspace{-0.0cm}
\centering
\includegraphics[width=1.0\textwidth]{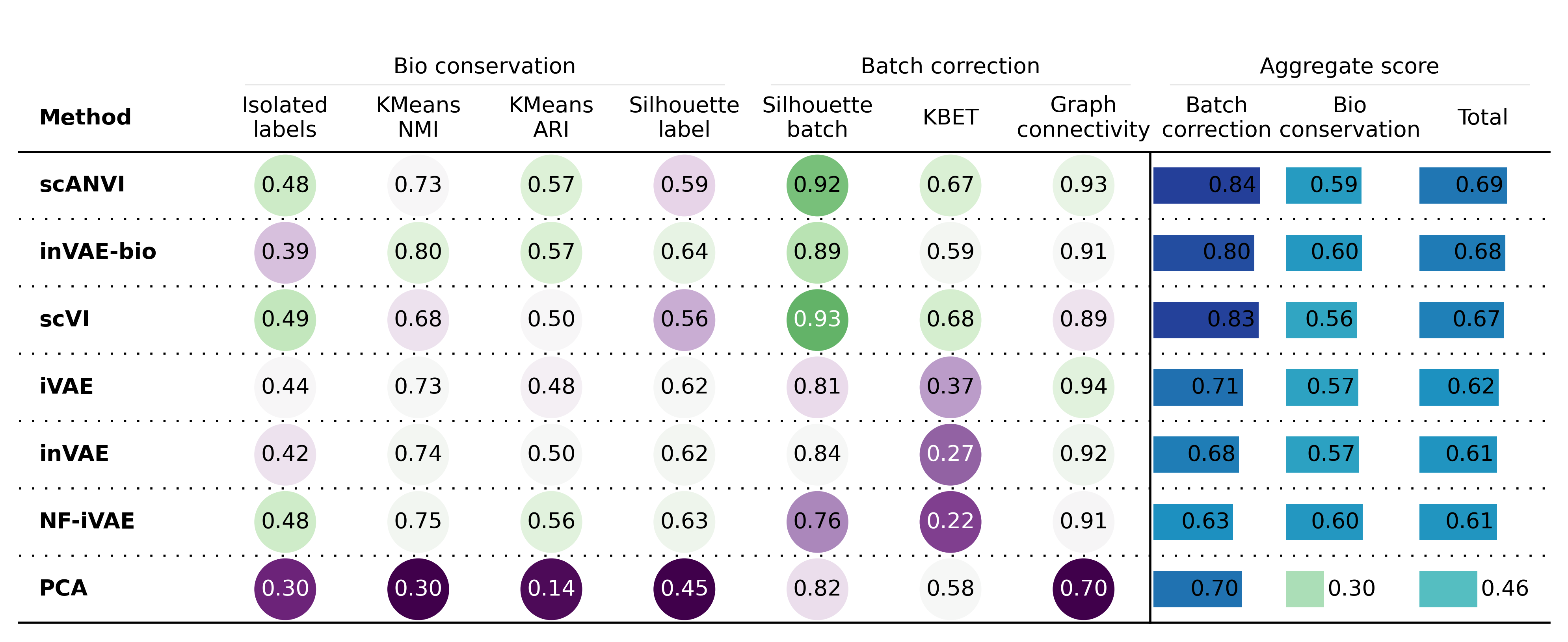}
\caption{Assessing the integration results across batches with similar donors.}\label{fig:scIB}
\end{figure}
To explore invariant representation learning, we initially focus on the human hematopoiesis data from~\cite{luecken2021a}. 
For training, we utilize three datasets consisting of 15,496 cells from two distinct donors, generated at two different sites, with 21 cell types (classes) (additional details in Appendix~\ref{app:sec:data}). 
We anticipate that datasets with similar donors but different sites would exhibit biological similarities, making the variability across those datasets undesirable. 
Conversely, we assume that datasets from different donors should contain interesting biological variations.

To visualize the data and compare the latent representations of different methods, we utilize UMAP visualizations~\cite{mcinnes2018uniform} colored with three batches, as illustrated in Figure~\ref{fig:motivation}.
Without any correction using the raw data (top left), the cells are clustered based on their batch, making it challenging to interpret the heterogeneity across cells.
For downstream tasks, we employ the invariant latent variables of inVAE and refer to it as inVAE-bio. 
While all other methods successfully cluster similar cells (Figure~\ref{fig:motivation} bottom right and Figure~\ref{fig:cell_types}), only inVAE-bio effectively separates batch s1d2 (donor 2) from s1d1 and s2d1 (donor 1) while simultaneously eliminating technical noise from samples with similar donors.
These observations are further evaluated in Figure~\ref{fig:scIB} using scIB metrics~\cite{luecken2022benchmarking,scIB}. 
scVI and scANVI aggressively mask batch effects, whereas inVAE-bio achieves higher biological conservation and interpretability by disentangling noise from biological signals. 
On the other hand, inVAE using both invariant and spurious latent variables, as well as iVAE and NF-iVAE, perform poorly, as expected due to the presence of spurious correlations in their representations.

\subsection{Predicting cell types using invariant predictors}
Classifying cell types (also called cell annotation) in new datasets poses a challenge in single-cell genomics due to technical variations and biases across datasets generated in different environments. 
We examine the effectiveness of various deep generative models for cell type classification. The prediction results are shown in Table~\ref{table:cell_annotation}. 
The results indicate that inVAE-bio surpasses the others and achieves the highest performance on two unseen datasets with different donors and sites (additional details in Appendix~\ref{app:sec:data}). 
The superior performance of inVAE, compared to inVAE-bio, on training datasets can be attributed to the presence of spurious correlations in its latent representation, leading to overfitting.

\begin{table}[t]
    \caption{Accuracy Results for cell annotation in human hematopoiesis data. 
    The table presents average, minimum, maximum, and median test results across 21 cell types.\label{table:cell_annotation}}
    \centering
        \begin{small}
            \begin{sc}
                \begin{tabular}{lcccccc}
                    \toprule
                     & & &\multicolumn{4}{c}{test (\%)} \\
                    \cmidrule(lr){4-7}
                    \multicolumn{1}{c}{model} & \multicolumn{1}{c}{train} &  \multicolumn{1}{c}{validation} & \multicolumn{1}{c}{avg} & \multicolumn{1}{c}{min} & \multicolumn{1}{c}{max} & \multicolumn{1}{c}{median}\\
                    \midrule

                    inVAE-bio & $0.891 \pm 0.001$ & $\mathbf{0.865 \pm 0.000}$ & $\mathbf{0.653 \pm 0.003}$ & $\mathbf{0.352}$ & $\mathbf{0.988}$ & $0.821$\\
                    scANVI & $0.818 \pm 0.015$ & $0.798 \pm 0.011$ & $0.579 \pm 0.028$ & $0.318$ & $0.977$ & $\mathbf{0.830}$\\
                    inVAE & $\mathbf{0.909 \pm 0.001}$ & $0.862 \pm 0.001$ & $0.557 \pm 0.004$ & $0.109$ & $0.951$ & $0.728$\\
                    iVAE & $0.818 \pm 0.003$ & $0.780 \pm 0.006$ & $0.563 \pm 0.017$ & $0.127$ & $0.979$ & $0.709$ \\
                    NF-iVAE & $0.809 \pm 0.001$ & $0.767 \pm 0.004$ & $0.507 \pm 0.004$ & $0.181$ & $0.914$ & $0.654$\\
                    \bottomrule
                \end{tabular}            
            \end{sc}
        \end{small}
\end{table}

\subsection{Invariant representation with interventions}
\begin{figure}[!t]
\centering
\includegraphics[width=1.0\textwidth]{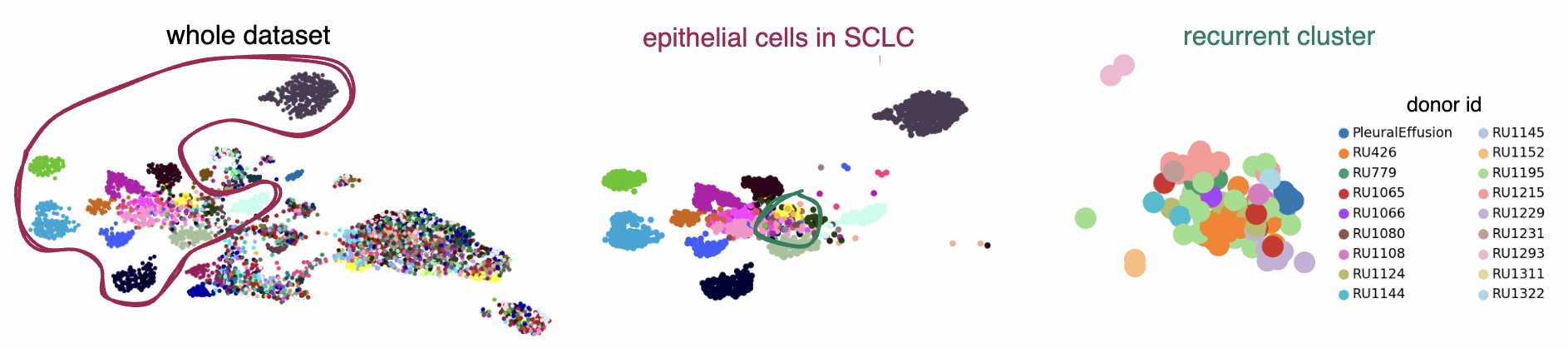}
\caption{inVAE-bio representation for lung cancer dataset. According to insights from the literature, the epithelial cells in SCLC exhibit variability among patients. Notably, a small subset of cells known as the recurrent cluster consists of epithelial cells that demonstrate similarity across all patients. }\label{fig:LC:ours:recurrent}
\end{figure}

Lastly, we examine the integration of human lung cancer data from~\cite{chan2021signatures}. 
The dataset comprises 155,098 cells from 21 Small Cell Lung Cancer (SCLC) clinical samples obtained from 19 patients, along with 24 Lung Adenocarcinoma (LUAD) and 4 tumor-adjacent normal lung samples serving as controls (details in~\ref{app:sec:data}).
According to literature~\cite{chan2021signatures}, we anticipate variability among epithelial cells in SCLC samples across different patients, while expecting minimal biological variations among other cells. 
UMAP visualizations of control and cancer cells using inVAE-bio are presented in Figure~\ref{fig:LC:ours:recurrent} and \ref{fig:LC:ours:covariates}. 
Our method effectively captures patient-level variability (Figure\ref{fig:LC:ours:recurrent}, middle) and removes spurious correlations in other cells (Figure~\ref{fig:LC:ours:recurrent}, left).
However, scVI (and scANVI) over-corrects the batch effect and fails to capture relevant biological variations across samples (Figure~\ref{fig:LC:scvi}). In~\cite{chan2021signatures}, a small subset of epithelial cells in SCLC samples is annotated as recurrent, containing samples from all batches. 
Our results in Figure~\ref{fig:LC:ours:recurrent} (right) demonstrate that our method successfully captures the similarity among these cells and clusters them appropriately.
\section{Conclusions}
In this study, we introduce inVAE, a deep generative model that utilizes identifiable variational autoencoders. 
Our model infers a two-partition latent space, where one part captures spurious correlations or biases that vary across domains, while the other remains invariant to unwanted variability. 
Importantly, the latent features in these partitions are independent, promoting the disentanglement of the invariant representation from the varying representation.
We apply our approach to large and heterogeneous single-cell gene expression data for cell state detection and cell type classification. 
Our results demonstrate that inVAE addresses a main limitation of current deep generative models in single-cell biology, namely their inability to disentangle biological signals from unrelated data biases during data integration. 
We envision that our model will find applications in more complex scenarios, such as cell state detection in multi-scale biological data (e.g., genes, proteins, chromatin accessibility, spatial gene expression), as well as in computer vision for tasks like style-content isolation in images.
It is important to take caution when making consequential decisions, particularly in healthcare, based on the structures learned from observational data.

\section*{Acknowledgment}
We express our gratitude to Stefan Bauer for engaging discussions on causal models and generative processes, as well as his feedback on the paper. 
We would also like to acknowledge Sergei Rybakov for reviewing the equations and providing valuable feedback. 
We extend our appreciation to Jason Hartford and Sebastien Lachapelle for fruitful discussions on identifiability during the initial phases of this project. 
Finally, we are thankful to Julien Gagneur and his administrative team, particularly Florian H\"olzlwimmer, for their generous and invaluable support in providing computational resources.

\bibliographystyle{unsrt}
\bibliography{main}


\newpage
\appendix
\section{Related work: Data Integration\label{app:sec:integration}}
Deep generative models are common frameworks for integration of single cell RNA sequencing data. Single cell Variational Inference (scVI)~\cite{lopez2018deep} appends one-hot encoding of the variables denoting the environments (e.g. experimental batches, sequencing protocols, site, etc) to genes which are then used as inputs to a VAE. 
The one-hot encoding is appended to the first layer of both the encoder and the decoder and has proven to be highly effective in removal of technical noise, also called batch effect, in certain datasets. 
In this case scVI is akin to conditional VAEs (cVAEs) \cite{sohn2015learning}. 
While scVI is completely unsupervised, scanVI \cite{xu2021probabilistic}, an extension of scVI,  is a semi-supervised deep generative model based on M1 + M2 model in~\cite{kingma2014semi} which additionally uses cell annotations (cell type labels) to integrate cells across environments. 

Both scVI and scANVI are widely recognized as state-of-the-art integration models in the single-cell community. 
The latent representations produced by these models primarily capture variations corresponding to cell types, which represent the primary axis of variability in single-cell data, disregarding technical noise. 
However, other sources of biological variation, including inter-patient differences and treatment effects (especially when the treatment's impact is subtle), tend to become entangled with batch effects and may be masked as a result. 
Consequently, these methods are susceptible to potential over-correction issues.

\section{ELBO for inVAE\label{app:sec:elbo:ours}}
  \begin{equation}
 \begin{split}
     \log p_\theta(\textsc{x}|\textsc{u}) \\
     =& \log p_\theta(\textsc{x}|\textsc{u})\int q_\varphi(\textsc{z}|\textsc{x},\textsc{u}) d\textsc{z} \\
     =& \int \log p_\theta(\textsc{x}|\textsc{u}) q_\varphi(\textsc{z}|\textsc{x},\textsc{u}) d\textsc{z}\\
     =& \mathbb{E}_{q_\varphi(\textsc{z}|\textsc{x},\textsc{u})} \left[\log p_\theta(\textsc{x}|\textsc{u})\right] \\
     =& \mathbb{E}_{q_\varphi(\textsc{z}|\textsc{x},\textsc{u})} \left[\log \frac{p_\theta(\textsc{x},\textsc{z}|\textsc{u})}{p_\theta(\textsc{z}|\textsc{x},\textsc{u})}\right]\\
     =& \mathbb{E}_{q_\varphi(\textsc{z}|\textsc{x},\textsc{u})} \left[\log \frac{p_\theta(\textsc{x},\textsc{z}|\textsc{u})q_\varphi(\textsc{z}|\textsc{x},\textsc{u})}{p_\theta(\textsc{z}|\textsc{x},\textsc{u})q_\varphi(\textsc{z}|\textsc{x},\textsc{u})} \right]\\
     =& \mathbb{E}_{q_\varphi(\textsc{z}|\textsc{x},\textsc{u})} \left[\log \frac{p_\theta(\textsc{x},\textsc{z}|\textsc{u})}{q_\varphi(\textsc{z}|\textsc{x},\textsc{u})} \right]+ D_\text{KL}\left( q_\varphi(\textsc{z}|\textsc{x},\textsc{u})||p_\theta(\textsc{z}|\textsc{x},\textsc{u})\right)\\
     \geq& \mathbb{E}_{q_\varphi(\textsc{z}|\textsc{x},\textsc{u})} \left[\log \frac{p_\theta(\textsc{x},\textsc{z}|\textsc{u})}{q_\varphi(\textsc{z}|\textsc{x},\textsc{u})}\right]\\
     =& \mathbb{E}_{q_\varphi(\textsc{z}|\textsc{x},\textsc{u})}\left[\log \frac{p_\phi(\textsc{x}|\textsc{z})p_\Theta(\textsc{z}|\textsc{u})}{q_\varphi(\textsc{z}|\textsc{x},\textsc{u})}\right]\\
     =& \mathbb{E}_{q_\varphi(\textsc{z}|\textsc{x},\textsc{u})}\left[\log \frac{p_\phi(\textsc{x}|\textsc{z})p_\Theta(\textsc{z}_I,\textsc{z}_S|\textsc{d},\textsc{e})}{q_\varphi(\textsc{z}|\textsc{x},\textsc{u})}\right]\\
     =& \mathbb{E}_{q_\varphi(\textsc{z}|\textsc{x},\textsc{u})}\left[\log \frac{p_\phi(\textsc{x}|\textsc{z})p_\Theta(\textsc{z}_I|\textsc{d})p_\Theta(\textsc{z}_S|\textsc{e})}{q_\varphi(\textsc{z}|\textsc{x},\textsc{u})}\right]\\
     =& \mathbb{E}_{q_\varphi(\textsc{z}|\textsc{x},\textsc{u})} \left[\log p_\phi(\textsc{x}|\textsc{z}) + \log \frac{p_\Theta(\textsc{z}_I|\textsc{d})p_\Theta(\textsc{z}_S|\textsc{e})}{q_\varphi(\textsc{z}|\textsc{x},\textsc{u})}\right]\\
     =& \mathbb{E}_{q_\varphi(\textsc{z}|\textsc{x},\textsc{u})}\left[\log p_\phi(\textsc{x}|\textsc{z})\right] - D_{KL} (q_\varphi(\textsc{z}|\textsc{x},\textsc{u})||p_\Theta(\textsc{z}_I|\textsc{d})p_\Theta(\textsc{z}_S|\textsc{e}))
  \end{split}
  \end{equation}
Given $\int q_\varphi(\textsc{z}|\textsc{x},\textsc{u}) d\textsc{z} = 1$.

\section{Score matching\label{app:sec:score}}
We provide here a short summary of the \emph{Score Matching} (SM) method by \cite{hyvarinen2005estimation}.

Let us assume that we observe data $\textsc{x} \sim p_\textsc{x}(\cdot)$ with $\textsc{x} \in \real^n$ where we want to approximate the observed density $p_x (\cdot)$ by some parametric distribution with density $p(\cdot ; \theta)$ with $\theta \in \real^t$. 
We assume that we can only calculate the density up to a multiplicative constant, i.e.
\begin{equation*}
p(\xi; \theta) = \frac{1}{\mathcal{Z}(\theta)} \tilde{p} (\xi; \theta)
\end{equation*}
where we know how to calculate the non-normalized density $\tilde{p}(\xi; \theta)$ (e.g. by an analytical form) but can not calculate the normalization constant $\mathcal{Z}(\theta)$, for example in the case where the integral $\int_{\xi \in \real^n} \tilde{p}(\xi; \theta) \, d \xi$ is intractable. 
Estimation by score matching now works by first defining the score function (in our context) to be the gradient of the log density with respect to the data vector:
\begin{equation}
\psi (\xi; \theta) = \begin{pmatrix}
\frac{\partial \log p(\xi; \theta)}{\partial \xi_1} \\
\vdots \\
\frac{\partial \log p(\xi; \theta)}{\partial \xi_n}
\end{pmatrix} =
\begin{pmatrix}
\psi_1 (\xi; \theta) \\
\vdots \\
\psi_n (\xi; \theta)
\end{pmatrix} = \nabla_\xi \log p(\xi;\theta).
\end{equation}
Here, it is important to note that the score function of our model (the parametric form of the density we chose) does not depend on $\mathcal{Z}(\theta)$, i.e. $\psi (\xi; \theta) = \nabla_\xi \log \tilde{p}(\xi; \theta)$. To go on, \cite{hyvarinen2005estimation} now define, together with the score function of the data $\psi_\textsc{x} (\cdot) = \nabla_\xi \log p_\textsc{x}(\cdot)$, the score matching estimator of $\theta$ by
\begin{equation}\label{eq:sm_est_abstract}
J (\theta) = \frac{1}{2} \int_{\xi \in \real^n} p_\textsc{x} (\xi) \norm{\psi(\xi;\theta) - \psi_\textsc{x} (\xi)}^2 \, d \xi
\end{equation}
and the corresponding estimate of $\theta$ is $\hat{\theta} = \argmin_\theta J(\theta)$. This theoretic estimator, that works by minimizing the expected squared distance between the model and the data score function, is justified by the following two points:
\begin{enumerate}
\item Although calculating the score function of the data $\psi_\textsc{x} (\cdot)$ is again a hard non-parametric estimation problem, we can use the score matching estimator with a different, equivalent minimization objective avoiding this computation (which is one of the main results in \cite{hyvarinen2005estimation}) to estimate the parameters $\theta$.

\item Even more important, under some conditions, minimizing the score matching estimator $J(\theta)$ corresponds to finding the true parameter $\theta^*$ such that $p_\textsc{x} (\cdot) = p(\cdot; \theta^*)$.
\end{enumerate}

For point 1 above, it is shown in \cite{hyvarinen2005estimation} that the score matching estimator \eqref{eq:sm_est_abstract}, assuming $\psi (\xi; \theta)$ is differentiable and some weak regularity conditions are fulfilled, can be also written as
\begin{equation}\label{eq:sm_estimator_abstract}
J(\theta) = \int_{\xi \in \real^n} p_\textsc{x} (\xi) \sum_{i=1}^n \left[ 
	\frac{\partial^2 \log \tilde{p}(\xi; \theta)}{\partial \xi_i^2} +
	\frac{1}{2} \left( \frac{\partial \log \tilde{p}(\xi; \theta)}{\partial \xi_i} \right)^2
\right] \, d \xi + const.
\end{equation}
where the constant does not depend on $\theta$. In practice, we can sample this estimator $J$ via $T$ observations of the random variable $\textsc{x}$, i.e. $(\textsc{x}^{(1)}, \ldots, \textsc{x}^{(T)})$, by
\begin{equation}
\tilde{J} (\theta) = \frac{1}{T} \sum_{t=1}^{T} \sum_{i=1}^{n} \left[ 
	\frac{\partial^2 \log \tilde{p}(\textsc{x}^{(t)}; \theta)}{\partial (\textsc{x}^{(t)}_i)^2} +
	\frac{1}{2} \left( \frac{\partial \log \tilde{p}(\textsc{x}^{(t)}; \theta)}{\partial \textsc{x}^{(t)}_i} \right)^2
\right]
\end{equation}
where we drop the constant for the purpose of optimizing $\tilde{J} (\theta)$ over $\theta$. 

Furthermore, many extensions of the score matching method exist. 
One of these extensions, that we call regularized score matching in the following, is given by \cite{lin2016estimation}. 
They argue that the conditions necessary for the above derivations to hold, are often violated in practice, especially the assumption of a smooth enough data density. 
Therefore, they derive the regularized score matching estimator by an approximation to the {true} SM estimator for input data with Gaussian noise as
\begin{equation}
\tilde{J} (\theta) = \frac{1}{T} \sum_{t=1}^{T} \sum_{i=1}^{n} \left[ 
	\frac{\partial^2 \log \tilde{p}(\textsc{x}^{(t)}; \theta)}{\partial (\textsc{x}^{(t)}_i)^2} +
	\frac{1}{2} \left( \frac{\partial \log \tilde{p}(\textsc{x}^{(t)}; \theta)}{\partial \textsc{x}^{(t)}_i} \right)^2 + \lambda_{regSM} \left( \frac{\partial^2 \log \tilde{p}(\textsc{x}^{(t)}; \theta)}{\partial (\textsc{x}^{(t)}_i)^2} \right)^2
\right], \label{eq:reg_sm_estimator}
\end{equation}
where $\lambda_{regSM}$ is a hyperparameter controlling the strength of the regularization. 
This hyperparameter can be thought of as smoothing the optimization landscape for score matching. 
In our experiments, we randomly sampled $\lambda_{regSM}$ as either zero, i.e. regular SM, or as a random non-zero value.
\section{Minibatch-weighted sampling\label{app:sec:minibatch}}

Here, we extend the proof in Appendix C of \cite{NEURIPS2018_1ee3dfcd}.
Let $\mathcal{B}^\textsc{u} = \{\textsc{x}_1,\dots,\textsc{x}_b\}^\textsc{u}$ be a minibatch of $b$ indices from domain $\textsc{u}$ where each element is sampled i.i.d from $p_{\textsc{x}|\textsc{u}}$. 
So, for any sampled batch instance  $\mathcal{B}^\textsc{u}$, $p(\mathcal{B}^\textsc{u})=(1/n^\textsc{u})^b$, where $n^\textsc{u}$ is the total number of samples from domain $\textsc{u}$.

\begin{align}
\mathbb{E}_{q(\textsc{z}|\textsc{u})}\left[\text{log}~ q(\textsc{z}|\textsc{u})\right] &=
\mathbb{E}_{q(\textsc{z},\textsc{x}_i|\textsc{u})}\left[\text{log}~\mathbb{E}_{\textsc{x}_j\sim p_{\textsc{x}|\textsc{u}}}\left[ q(\textsc{z}|\textsc{x}_j,\textsc{u})\right]\right] \\
&= \mathbb{E}_{q(\textsc{z},\textsc{x}_i|\textsc{u})}\left[\text{log}~\mathbb{E}_{p(\mathcal{B}^\textsc{u})}\left[ \frac{1}{b}\sum_{j=1}^{b}q(\textsc{z}|\textsc{x}_j,\textsc{u})\right]\right] \\
&\geq \mathbb{E}_{q(\textsc{z},\textsc{x}_i|\textsc{u})}\left[\text{log}~\mathbb{E}_{p(\mathcal{B}^\textsc{u}|\textsc{x}_i)}\left[ \frac{p(\mathcal{B}^\textsc{u})}{p(\mathcal{B}^\textsc{u}|\textsc{x}_i)}\frac{1}{b}\sum^{j=1}_{b}q(\textsc{z}|\textsc{x}_j,\textsc{u})\right]\right] \\
&= \mathbb{E}_{q(\textsc{z},\textsc{x}_i|\textsc{u})}\left[\text{log}~\mathbb{E}_{p(\mathcal{B}^\textsc{u}|\textsc{x}_i)}\left[ \frac{1}{n^{\textsc{u}}b}\sum_{i=1}^{b}q(\textsc{z}|\textsc{x}_j,\textsc{u})\right]\right]\
\end{align}
 This can be estimated as follows (more discussions in~\cite{NEURIPS2018_1ee3dfcd}):
\begin{align}
\mathbb{E}_{q(\textsc{z}|\textsc{u})}\left[\text{log}~ q(\textsc{z}|\textsc{u})\right] &\approx 
    \frac{1}{b}\sum_{i=1}^{b} \left[\text{log}~\frac{1}{n^\textsc{u}b}\sum_{j=1}^{b}q(\textsc{z}(\textsc{x}_i)|\textsc{x}_j,\textsc{u})\right]
\end{align}
where $p(\mathcal{B}^\textsc{u}|\textsc{x}_i)$ is the probability of the sampled minibatch where one of its element is fixed to be $\textsc{x}_i$ and the rest are sampled i.i.d from $p_{\textsc{x}|\textsc{u}}$.
This can be intuitively interpreted as having a mixture distribution over each domain $\textsc{u}$, where the data index $j$ indicates a mixture component.
\section{The training loss in Practice\label{app:sec:training_loss}}
In practice, we follow \cite{lu2021nonlinear} and model only the log of the unnormalized probability of our invariant prior, i.e. the part inside the exponential function in Eq.~\eqref{eq:NF_prior}. This is done in the following way
\begin{align*}
\log \tilde{p}_{T,\lambda}(\textsc{z}_I|\textsc{d})
&= \left< T_{NN} (\textsc{z}_I), \lambda_{NN} (\textsc{d}) \right> + \left<T_f(\textsc{z}_I), \lambda_f (\textsc{d}) \right> \\
&= \left<NN(\textsc{z}_I;param1), NN(\textsc{d}; param2) \right> + \left<concat(\textsc{z}_I,\textsc{z}^2_I), NN(\textsc{d};param3) \right>
\end{align*}
where $\left< \cdot,\cdot \right>$ is the dot product of two vectors and $concat(\cdot,\cdot)$ denotes the concatenation of two vectors. 
Here, every part except $T_f(\textsc{z}_I)$ is implemented through their own respective neural network with corresponding input variables and correct output dimension. 
For the output dimension we only need to make sure that the respective outputs of the dot product have the same dimensions, but for example $T_{NN} (\textsc{z}_I)$ and $T_f (\textsc{z}_I)$ do not need to have the same size. 
The first dot product allows the prior to capture, as previously mentioned, arbitrary dependencies between the latent variables since $NN(\textsc{z}_I;param1)$ can model complicated nonlinear transformations of the latents. 
Going on, $T_f (\textsc{z}_I)$ is the concatenation of the latent variables and their square values and therefore fulfills the theoretic requirement of the definition \ref{eq:NF_prior} of the prior that each $T_i (\textsc{z}_i)$ has at least 2 dimensions. 
Finally, the second dot product (without the first dot product) corresponds to a \emph{factorized} exponential family.

Concretely, the training loss in Eq.~\ref{eq:training_loss}, 
is optimized as follows. 
The $\mathcal{L}_{ELBO}$ part of the loss is responsible for optimizing the parameters of the encoder $q_\varphi(\textsc{z}|\textsc{x},\textsc{u})$, decoder $p_\phi(\textsc{x}|\textsc{z})$ and {spurious} prior $p_\Theta(\textsc{z}_S|\textsc{e})$ while keeping the invariant prior $\tilde{p}_\Theta(\textsc{z}_I|\textsc{d})$ as fixed. 
In practice, this is done via the \emph{Pytorch} framework by setting \emph{requires\_grad} to False for the respective operations of the invariant prior for the calculation of $\mathcal{L}_{ELBO}$ such that its parameters are not updated through that part. 
Similarly, the score matching part in the loss $\mathcal{L}_{SM}$ only updates the parameters of the invariant prior $\tilde{p}_\Theta(\textsc{z}_I|\textsc{d})$ by keeping the rest of the model as fixed. 
This is achieved by, first, calling \emph{detach} on the latent space variable $\textsc{z}$, so that the parameters of the encoder are not updated by the score matching loss. 
Secondly, also keeping the parameters of the spurious prior fixed, by either setting \emph{requires\_grad} to False or not calculating the gradients in the score matching loss with respect to the indices that belong to the spurious prior.

\section{Identifiability Proof\label{app:sec:identifiability}}

\noindent\textbf{Identifiability of the invariant partition of the latent representation}

\noindent We place a non-factorized prior $p_{T,\lambda}(\textsc{z}_I|\textsc{d})$ from the exponential family distribution \cite{lu2021nonlinear} on the invariant partition $\textsc{z}_I$, where $\textsc{z}_I \in \real^{i}$, is shown to be identifiable under certain conditions based on the results in~\cite{lu2021nonlinear}. The identifiability of latent representations belonging to this partition imply that the $\sim_A$- and $\sim_P$-equivalence relations (see Definition 8 and 9 in \cite{lu2021nonlinear}), that is identifiability up to simple transformations and permutations respectively, hold. Therefore:

\begin{equation}\label{eq:class_a_invariant}
    \exists A_I, c_I \text{ s.t. } T(f^{-1}(\textsc{z}_I)) = A_I  T'(f'^{-1}(\textsc{z}_I)) + c_I, 
\end{equation}
where $A_I \in \real^{i \times i}$ is an invertible matrix and $c_I \in \real^{i}$ is a constant, and $T'$ and $f'$ are different parameterization of the model.
Similarly, there exists a block permutation matrix $P\in \real^{i \times i}$ such that: 
\begin{equation}\label{eq:class_p_invariant}
    \exists P_I, c_I \text{ s.t. } T(f^{-1}(\textsc{z}_I)) = P_I  T'(f'^{-1}(\textsc{z}_I)) + c_I, 
\end{equation}

\noindent\textbf{Identifiability of the spurious partition of the latent representation}

\noindent The prior on the spurious partition, $p_{T,\lambda}(\textsc{z}_S|\textsc{e})$, is designed to belong to a factorized exponential family distribution \cite{khemakhem2020variational},  where $\textsc{z}_S \in \real^{s}$, is proved to be identifiable under certain conditions as described in~\cite{khemakhem2020variational}. Therefore $\sim_A$- and $\sim_P$-equivalence relations also hold here and we have:
\begin{equation}\label{eq:class_a_sp}
    \exists A_S, c_S \text{ s.t. } T(f^{-1}(\textsc{z}_S)) = A_S  T'(f'^{-1}(\textsc{z}_S)) + c_S, 
\end{equation}
where $A_S \in \real^{s \times s}$ is an invertible matrix and $c_S \in \real^{s}$ is constant.
Similarly, there exists a block permutation matrix $P \in \real^{s \times s}$ such that: 
\begin{equation}\label{eq:class_p_sp}
    \exists P_S, c_S \text{ s.t. } T(f^{-1}(\textsc{z}_S)) = P_S  T'(f'^{-1}(\textsc{z}_S)) + c_S, 
\end{equation}

\noindent\textbf{Identifiability of joint latent representation}

\noindent To prove identifiability of the joint latent representation, we first demonstrate that the two priors can be viewed as a joint prior that belongs to a general exponential family distribution. 
We then use the assumptions and results from NF-iVAE in~\cite{lu2021nonlinear} to complete the proof.

Assumption 1 in~\cite{lu2021nonlinear} on causal graph is satisfied for our method as discussed in Section~\ref{sec:A1}. 
We assume that data is sampled from a deep generative model with parameters $\mathbb\theta = (f, T, \lambda
)$ defined according to:
\[p_\theta (\textsc{x},\textsc{z} | \textsc{d},\textsc{e}) = p_f (\textsc{x}|\textsc{z}) p_{T,\lambda} (\textsc{z}|\textsc{d},\textsc{e})\]
\[p_f (\textsc{x}|\textsc{z}) = p_\epsilon(\textsc{x} - f(\textsc{z}))\]
which means that the value of $\textsc{x}$ can be decomposed as $\textsc{x}= f(\textsc{z}) + \epsilon$, where $\epsilon$ is an independent noise variable with density function $p_\epsilon(\epsilon)$, and is independent of $\textsc{z}$ or $f$ (we show later that this assumption might not hold in real applications). 
The prior on the latent variables is represented as:

\begin{align}
    p_{T,\lambda} (\textsc{z}|\textsc{d},\textsc{e}) = p_{T,\lambda}(\textsc{z}_I, \textsc{z}_S | \textsc{d},\textsc{e}) &= p_{T,\lambda}(\textsc{z}_I|\textsc{d})p_{T,\lambda}(\textsc{z}_S|\textsc{e}) \\
    &= \frac{\mathcal{Q}(\textsc{z}_I)}{\mathcal{Z}(\textsc{d})}\exp\left[ T(\textsc{z}_I)^T\lambda(\textsc{d})\right] \Pi_{i=1}^{s}\frac{\mathcal{Q}_i(\textsc{z}_{S_i})}{\mathcal{Z}_i(\textsc{e})}\exp\left[ \sum_{j=1}^{k}T_{i,j}(\textsc{z}_{S_i})\lambda_{i,j}(\textsc{e})\right] \\
    &= \frac{\mathcal{Q}(\textsc{z})}{\mathcal{Z}(\textsc{d},\textsc{e})}\exp{\left[\underbrace{<T_f(\textsc{z}_I),\lambda_f(\textsc{d})> + <T_{NN}(\textsc{z}_I),\lambda_{NN}(\textsc{d})>}_{\mathclap{\text{non-factorized prior}}} + \underbrace{<T_f(\textsc{z}_S), \lambda_f(\textsc{e})>}_{\mathclap{\text{factorized prior}}}\right]}
\end{align}

If we concatenate $\textsc{z}_I$ and $\textsc{z}_S$ and their sufficient statistics $T$, the expression above can be written as:

\begin{align}
    p_{T,\lambda} (\textsc{z}|\textsc{d},\textsc{e}) 
    &= \frac{\mathcal{Q}(\textsc{z})}{\mathcal{Z}(\textsc{d},\textsc{e})}\exp{\left[<T_f(\textsc{z}),\lambda_f(\textsc{d},\textsc{e})> + <T_{NN}(\textsc{z}),\lambda_{NN}(\textsc{d},\textsc{e})>\right]} \\
    &= \frac{\mathcal{Q}(\textsc{z})}{\mathcal{Z}(\textsc{d},\textsc{e})}\exp{\left[T(\textsc{z})^T\lambda(\textsc{d},\textsc{e})\right]} \numberthis \label{eq:inVAE_def}
\end{align}

The resulting density belongs to a general exponential family with parameter vector given by arbitrary function $\lambda(\textsc{d},\textsc{e})$ and sufficient statistics $T(\textsc{z}) = \left[T_f(\textsc{z})^T, T_{NN}(\textsc{z})^T\right]^T$ given by concatenation of sufficient statistics of a factorised exponential family $T_f(\textsc{z})$ where all $T_i(\textsc{z}_i)$ have dimension larger or equal to 2, and $T_{NN}(\textsc{z})$ which is an output of a neural network with ReLU activations, as required by Assumption 2 in~\cite{lu2021nonlinear}.
We can intuitively view $T_{NN}(\textsc{z})$ as $T_{NN}((\textsc{Z}_I,\Vec{0}))$.

Additionally, the results in (\ref{eq:class_a_invariant}) and (\ref{eq:class_a_sp}) imply that there exists a block-diagonal, invertible matrix $A \in \real^{m \times m}$, with $m = i + s$, such that the parameter set $(f, T, \lambda
)$ is $\sim_A$-identifiable: 
\begin{equation}\label{eq:class_a}
    \exists A, c \text{ s.t. } T(f^{-1}(\textsc{z})) = A T'(f'^{-1}(\textsc{z})) + c, 
\end{equation}
where $c \in \real^{m}$ is a constant and the blocks in A correspond to $A_I$ and $A_S$. 
Since both $A_I$ and $A_S$ are invertible, the resulting block-diagonal matrix $A$ is also invertible. 
Thus, the $\sim_A$ equivalence relation still holds for $\mathbb\theta = (f, T, \lambda)$ given the results from iVAE \cite{khemakhem2020variational} and NF-iVAE  \cite{lu2021nonlinear} and independence of $\textsc{z}_I$ and $\textsc{z}_S$. 

A similar reasoning can be argued to demonstrate $\sim_P$ equivalence relation under certain conditions on parameters holds for a block permutation matrix $P$, $m \times m$, using the results in (\ref{eq:class_p_invariant}) and (\ref{eq:class_p_sp}) and independence of $\textsc{z}_I$ and $\textsc{z}_S$. 
Hence, the parameter set $(f, T, \lambda)$ is also $\sim_P$-identifiable:

\begin{equation}\label{eq:class_p}
    \exists P, c \text{ s.t. } T(f^{-1}(\textsc{z})) = P  T'(f'^{-1}(\textsc{z})) + c, 
\end{equation}
where $c \in \real^{m}$ is constant and $T'$ and $f'$ are different parameterization of model parameters $T$ and $f$.

The parameters $\theta$ are therefore identifiable up to a simple transformation and a permutation of the latent variables $\textsc{z}$. 
That is, for all parameter sets $\theta = (f, T, \lambda
)$ and $\theta' = (f', T', \lambda')$, we have:
\[\forall(\theta,\theta'): p_\textsc{x}(\theta) = p_\textsc{x}(\theta') \Rightarrow \theta \sim \theta' \]   

\noindent\textbf{Notes on identifiability assumptions in single-cell models}

\noindent This work and similar works in the field \cite{khemakhem2020variational, lachapelle2022disentanglement, lopez2022learning} assume that $\textsc{x}$ can be decomposed as $\textsc{x} = f(\textsc{z}) + \epsilon$, where $f$ is a differentiable bijection with a differentiable inverse. 
This is in general a very strong assumption in single-cell RNAseq, where the generative models estimate the parameters of the data distribution from $f(\textsc{z})$ to model gene expression count data \textsc{x}. 
So, the assumptions made on the decoder $f$ in the previous works where theoretical proofs of identifiabilty are presented may not hold in discrete gene expression count data, as raised and discussed in \cite{lopez2022learning}. 
However, empirical observations in \cite{khemakhem2020variational,lopez2022learning}  demonstrate improved performance for data with discrete features. 
Additionally, there are concerns regarding the size effect of $\textsc{d}$ on the variability captured by the latent features $\textsc{z}_I$, that is how the conditional distribution $p(\textsc{z}_I|\textsc{d})$ sufficiently vary with $\textsc{d}$ \cite{brehmer2022weakly, lopez2022learning}. 
We leave closer investigations of the validity of the above assumption on the decoder as well as the evaluation of the sufficient variability as future directions.

\section{Sparse mechanism shift}

To represent cellular perturbations as interventions on latent variables, related work assumes that the underlying mechanisms undergo sparse shifts when subjected to interventions~\cite{lopez2022learning}. In the previous method, called sVAE+~\cite{lopez2022learning}, an extension of sVAE \cite{lachapelle2022disentanglement} to single-cell data, each perturbation is considered a stochastic intervention that targets an unknown, yet sparse, subset of latent variables. Specifically, only a few latent components $\textsc{z}_i$ are influenced by perturbation $\textsc{p}$, while the other components remain invariant. This can be expressed as follows:

\begin{align}
\textsc{z}_i|\textsc{p} \sim \gamma_i^\textsc{p}N(f(\textsc{p}),1) + (1-\gamma_i^\textsc{p}) N(0,1)
\end{align}

Here, $\gamma_i^\textsc{p}$ is a learnable binary variable assumed to be sparse, and $N(f(\textsc{p}),1)$ represents a normal distribution with a mean determined by the function $f(\textsc{p})$, which depends on the perturbation $\textsc{p}$.
The extension of this work to non-factorized priors, where the latent components can be dependent, is an interesting follow-up research direction that we defer to future work.
\section{Architectures\label{app:sec:arch}}
All the neural networks used in this work are fully connected, feed-forward neural networks.
For our experiments, we sampled randomly from reasonable ranges for the relevant hyperparameters and chose the setting with the best ELBO on the validation set. In general, we found that usual settings for VAEs, for example encoder and decoder as a neural network with 2 hidden layers and 128  neurons with ReLU activation functions, work well for our models.
Furthermore, for a fair comparison, all the methods in Figure~\ref{fig:scIB} including iVAE and NF-iVAE are extended to use Negative-Binomial distribution for modeling the likelihood of the gene expression counts (i.e. for the decoder). 

In practice, for inVAE we use the following distributions: Gaussian with diagonal covariance for the encoder, Negative-Binomial for the decoder, the non-factorized general exponential family distribution for the invariant prior and a Gaussian distribution with diagonal covariance for the spurious prior. For inVAE, as well as iVAE and NF-iVAE, the gene expression counts are fed to the encoder by taking the $\log (\textsc{x} + 1)$ for numerical stability, same as per default for scVI. Additionally, we use Batch-Norm layers in the encoder and decoder for the same reasons.

All models are optimized using Adam~\cite{kingma2017adam} 
with an initial learning rate of $0.01$.
We use PyTorch's default weight initialization scheme for the weights.
All models can be trained entirely on CPUs on consumer grade Laptop machines within minutes or hours (see table~\ref{table:exec_times}).
For training human hematopoiesis data, "site" is considered as a technical covariate and "donor" and "cell type" as biological covariates. 
For the lung cancer data, "assay" and "treatment" are used as technical covariates and "disease" and "cell type" as biological covariates. 

\begin{table}[t]
    \caption{Execution times per epoch for training (on CPU AMD Ryzen 7 with 1.70 GHz and 32GB RAM).\label{table:exec_times}}
    \centering
        \begin{small}
            \begin{sc}
                \begin{tabular}{lc}
                    \toprule
                    \multicolumn{1}{c}{model} & \multicolumn{1}{c}{Execution time (s)} \\
                    \midrule

                    inVAE & $4.87 \pm 0.29$ \\
                    scANVI & $9.04 \pm 0.30$ \\
                    scVI & $\mathbf{3.86 \pm 0.16}$ \\
                    iVAE & $4.76 \pm 0.27$ \\
                    NF-iVAE & $4.89 \pm 0.31$ \\
                    \bottomrule
                \end{tabular}            
            \end{sc}
        \end{small}
\end{table}

\section{Evaluation metrics\label{app:sec:evaluation}}

We quantified the quality of the data integration using the following metrics from \cite{luecken2022benchmarking}, which are designated metrics for the evaluation of integration of single cell data, implemented in the \textit{scIB} package. We provide a short explanation of each metric, for more details we suggest to read the original publication \cite{luecken2022benchmarking}.
Cell type ASW, isolated label F1, isolated label silhouette, NMI and ARI were used as biological conservation metrics. To quantify batch mixing we used principal component regression, graph connectivity and batch ASW.

\textit{NMI:} NMI (Normalized Mutual Information) measures the overlap between two different clusterings. This score is used between the cell type labels and the clusters obtained via unsupervised clustering of the dataset after integration. The score is bounded between 0 and 1, with 1 denoting a perfect match between the two clusterings and 0 the absence of any overlap.\\

\textit{ARI:} ARI stands for Adjusted Rand Index. The raw Rand Index scores the similarity between two clusterings, it considers both correct overlaps and correct disagreements in the computation. This score is computed between the cell type labels and the clusters obtained on the integrated dataset. The score spans from 0 to 1, with 1 representing a perfect score.
\\

\textit{KBET:} The kBET algorithm \cite{buttner2019test} determines whether the label composition of a k nearest neighborhood of a cell is similar to the expected (global) label composition. The test is repeated for a random subset of cells, and the results are summarized as a rejection rate over all tested neighborhoods. kBET works on a kNN graph.
\\

\textit{Cell type ASW:} ASW (Average Silhouette Width) is a measure of the relationship between the within-cluster distances of a cell and the between-cluster distances of the same cell to the closest cluster. This metric can range between -1 and 1. Values of -1 indicate total misclassification, values of 0 imply overlap between clusters and values of 1 occur when clusters are well separated. We use two versions of this scores, one computed on cell type labels, and a modified version to quantify batch mixing (see ASW batch below). The score is scaled to have values between 0 and 1 using the following equation:
\begin{equation*}
    cell type ASW = \frac{ASW_{C} + 1}{2},
\end{equation*}

where C represents the set of all cell type labels.
\\

\textit{Batch ASW:} 
The batch ASW quantifies the quality of batch mixing in the integrated object. We obtain it by computing the ASW but on batch labels instead of cell type labels. Scores of 0 are indicative of good batch mixing, while any deviation from this score is the result of batch effects. In order to have a metric bound between 0 and 1, the following transformation is applied:

\begin{equation*}
    batch ASW_{j} = \frac{1}{C_{j}}\sum_{i\in C_{j}} 1 - s_{batch}(i)
\end{equation*}

Here $C_{j}$ is the set of cells with label $j$ and $|C_{j}|$ is the support of the set.
The final score is obtained by averaging the batch ASW values obtained for each label:
\begin{equation*}
    batch ASW = \frac{1}{|M|} \sum_{j \in M} batch ASW_{j}
\end{equation*}
with M being the set of unique cell labels.

\textit{Isolated label F1:} Isolated labels are defined as cell type labels which are found in the smallest number of batches. We aim to determine how well these cell types are separated from the rest in the integrated data. To do so, we find the cluster containing the highest amount of cells from such isolated labels, and we then compute the $F_{1}$ score of the isolated label against all other labels within the cluster. We use the standard $F_{1}$ formulation:

\[
    F_{1} = 2 \frac{precision \cdot recall}{precision + recall}
\]
Once again, scores of 1 represent the desirable outcome in which all cells from the isolated labels are grouped in one cluster.

\textit{Isolated label silhouette:}
This is a cell type ASW score but computed only on the isolated labels.



\textit{Graph connectivity} This metric measures whether a kNN graph ($G$) computed on the integrated object connects cells that fall within the same cell type. It is bound between 0 and 1, with 0 indicating a graph where all cells are unconnected and, and 1 occurring when all cells of the same cell type are connected in the integrated output.
For each cell type label a subset graph $G(N_{c}, E_{c})$ is computed, the final score is obtained using the following formulation:
\begin{equation*}
    GC = \frac{1}{|C|} \sum_{c \in C} \frac{|LCC (G(N_{c}, E_{c}))|}{|N_{c}|}
\end{equation*}
where $C$ is the set of unique cell type labels, $|LCC()|$ is the number of nodes in the largest connected component in the kNN graph and $|N_{c}|$ is the number of nodes in the graph.
  
\section{Single-cell datasets\label{app:sec:data}}

\subsection{Human hematopoiesis}

This dataset contains 120,000 single cells from the human bone marrow of 5 diverse donors measured with two multi-modal technologies capturing gene expression profiles. This dataset is multi-site. Donors are all healthy
non-smokers without recent medical treatment. Donors varied by age (22 - 40), sex, and ethnicity. We were interested in preserving between-donor variation. We therefore use site as a distractor for the spurious prior.

Raw gene counts were downloaded as per author instructions \cite{luecken2021a}. For data processing, we follow the Best Practices guidelines for single cell genomics data analysis \cite{heumos2023best}. We select 2000 highly variable genes (HVGs) using the scanpy framework \cite{wolf2018scanpy} for processing of single cell data. 
We obtain UMAP visualisations using the scanpy pipeline 

The batch correction metrics often evaluate how well the cells from different batches are mixed within a neighborhood of cells. 
Therefore, the results in Figure~\ref{fig:scIB} are shown for the two training datasets with similar donors but different sites where we expect similar cells from different batches to cluster together.

\subsection{Human lung cancer}

This dataset contains 155,098 cells from 21 fresh Small Cell Lung Cancer (SCLC) clinical samples obtained from 19 patients, as well as 24 Lung Adenocarcinoma (LUAD) and 4 tumor-adjacent normal lung samples as controls. The SCLC and LUAD cohorts include treated and untreated patients. Samples were obtained from primary tumors, regional lymph node metastases, and distant metastases (liver, adrenal gland, axilla, and pleural effusion).

The data was downloaded from the Human Tumor Atlas portal, \url{data.humantumoratlas.org}. 
For data processing, we follow the Best Practices guidelines for single cell genomics data analysis \cite{heumos2023best}. 

\section{Extended results \label{app:sec:results}}
\subsection{Human hematopoiesis data\label{app:sec:results:hema}}

\begin{figure}[t]
\centering
\includegraphics[width=1.1\textwidth]{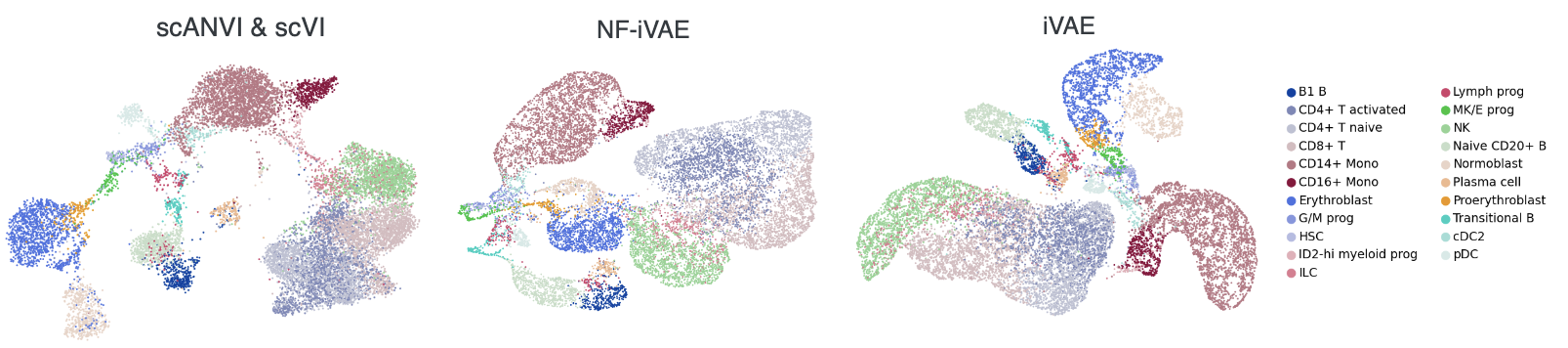}
\caption{The integration of human hematopoiesis datasets colored by cell types.}\label{fig:cell_types}
\end{figure}

The latent representations in Figure~\ref{fig:motivation} for scANVI, NF-iVAE, and iVAE colored by cell types are presented in Figure~\ref{fig:cell_types}.

\subsection{Lung cancer data\label{app:sec:results:lung}}

\begin{figure}[t]
\centering
\includegraphics[width=1.1\textwidth]{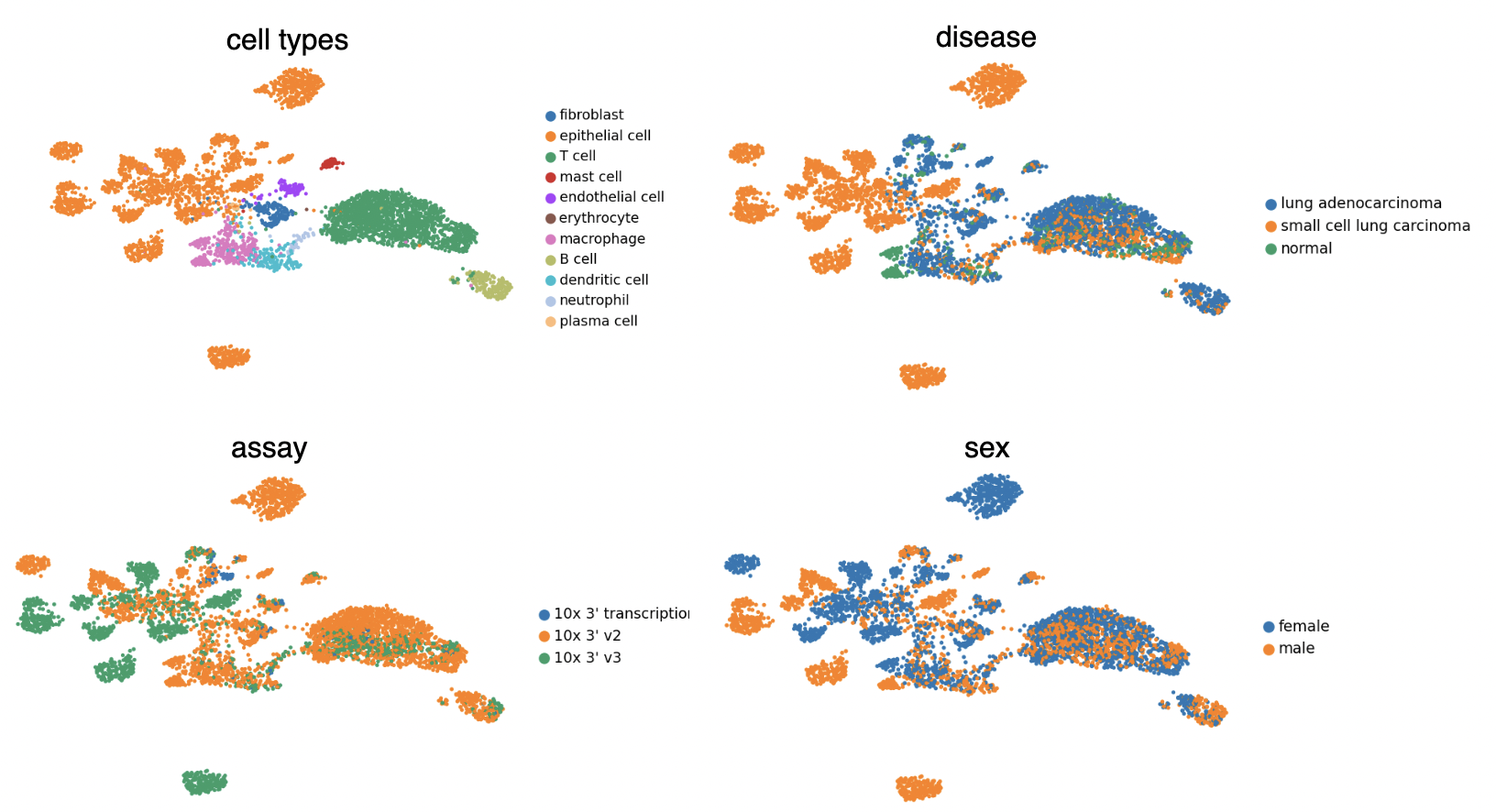}
\caption{The integration result of lung cancer dataset using inVAE colored by various covariates.}\label{fig:LC:ours:covariates}
\end{figure}

\begin{figure}[t]
\centering
\includegraphics[width=1.1\textwidth]{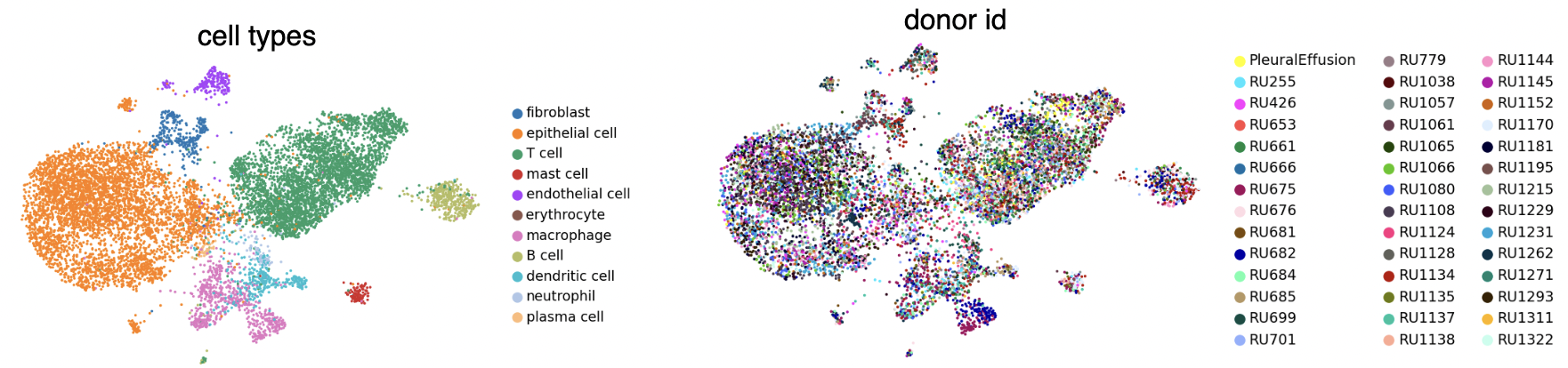}
\caption{The integration result of lung cancer dataset using scVI colored by batches. scVI over corrects the batch effect that results in masking biological variations across patients with SCLC.}\label{fig:LC:scvi}
\end{figure}

The data integration results using inVAE and scVI are presented in Figures~\ref{fig:LC:ours:covariates} and \ref{fig:LC:scvi}.

\end{document}

%% file: macros.tex
\DeclareMathOperator*{\argmin}{arg\,min}

\newcommand{\addref}[1]{{\color{Orange}[ref]}}

\theoremstyle{plain}

\theoremstyle{definition}

\newtheorem{ass}{Assumption}